\documentclass[a4paper,fleqn]{cas-sc}

\usepackage[authoryear]{natbib}
\usepackage{subcaption}
\usepackage{tcolorbox}
\usepackage{hyperref}
\usepackage{tablefootnote}
\usepackage{placeins}

\def\tsc#1{\csdef{#1}{\textsc{\lowercase{#1}}\xspace}}
\tsc{WGM}
\tsc{QE}
\tsc{EP}
\tsc{PMS}
\tsc{BEC}
\tsc{DE}

\begin{document}
\let\WriteBookmarks\relax
\def\floatpagepagefraction{1}
\def\textpagefraction{.001}

\shorttitle{DataFactory: Multi-Agent Framework for TableQA}

\shortauthors{Tong Wang, Chi Jin et~al.}

\title[mode = title]{DataFactory: Collaborative Multi-Agent Framework for Advanced Table Question Answering}                      
\tnotemark[1]\tnotemark[2]

\tnotetext[1]{This work was supported by the National Defense Science and Technology Key Laboratory Fund of China (Grant No. 6142006220506).}

\tnotetext[2]{This manuscript is the accepted version of the article published in \textit{Information Processing \& Management}. \copyright~2026 Elsevier Ltd. All rights reserved.}

\author[1]{Tong Wang}[%
   orcid=0000-0001-6981-916X
   ]
\fnmark[1]

\ead{tongwss@foxmail.com}

\credit{Conceptualization, Methodology, Software, Investigation, Writing - Original draft preparation}

\affiliation[1]{organization={Institute of Systems Engineering, Academy of Military Sciences},
    city={Beijing},
    postcode={100101}, 
    country={China}}

\author[2]{Chi Jin}[%
   orcid=0009-0009-4010-2718
   ]
\fnmark[1]

\ead{jinchi997@163.com}

\credit{Data curation, Investigation, Writing - Original draft preparation, Visualization}

\affiliation[2]{organization={School of Information Resource Management, Renmin University of China},
    city={Beijing},
    postcode={100872}, 
    country={China}}

\author[1]{Yongkang Chen}
\ead{yk.chen365@outlook.com}

\credit{Software, Validation, Formal analysis}

\author[1]{Huan Deng}
\ead{DHuan56@foxmail.com}

\credit{Software, Data curation, Resources}

\author[1]{Xiaohui Kuang}
\ead{xhkuang@bupt.edu.cn}

\credit{Project administration, Supervision, Writing - Review \& editing}

\author[1]{Gang Zhao}[%
   orcid=0009-0006-8668-777X
   ]
\cormark[1]
\ead{bisezhaog@163.com}

\credit{Conceptualization, Project administration, Supervision, Writing - Review \& editing}

\cortext[cor1]{Corresponding author}

\fntext[fn1]{These authors contributed equally to this work.}

\begin{abstract}
Table Question Answering (TableQA) enables natural language interaction with structured tabular data. However, existing large language model (LLM) approaches face critical limitations: context length constraints that restrict data handling capabilities, hallucination issues that compromise answer reliability, and single-agent architectures that struggle with complex reasoning scenarios involving semantic relationships and multi-hop logic. This paper introduces DataFactory, a multi-agent framework that addresses these limitations through specialized team coordination and automated knowledge transformation. The framework comprises a Data Leader employing the ReAct paradigm for reasoning orchestration, together with dedicated Database and Knowledge Graph teams, enabling the systematic decomposition of complex queries into structured and relational reasoning tasks. We formalize automated data-to-knowledge graph transformation via the mapping function $\mathcal{T}: \mathcal{D} \times \mathcal{S} \times \mathcal{R} \rightarrow \mathcal{G}$, and implement natural language-based consultation that—unlike fixed workflow multi-agent systems—enables flexible inter-agent deliberation and adaptive planning to improve coordination robustness; we also apply context engineering strategies that integrate historical patterns and domain knowledge to reduce hallucinations and improve query accuracy. Across TabFact, WikiTableQuestions, and FeTaQA, using eight LLMs from five providers, results show consistent gains. Our approach improves accuracy by 20.2\% (TabFact) and 23.9\% (WikiTQ) over baselines, with significant effects (Cohen's d>1). Team coordination also outperforms single-team variants (+5.5\% TabFact, +14.4\% WikiTQ, +17.1\% FeTaQA ROUGE-2). The framework offers design guidelines for multi-agent collaboration and a practical platform for enterprise data analysis through integrated structured querying and graph-based knowledge representation.
\end{abstract}


\begin{highlights}
\item Multi-agent framework transforms TableQA through specialized team coordination
\item Automated data-to-KG transformation enables multi-hop reasoning via mapping function
\item Database and KG teams achieve 20.2\% and 23.9\% accuracy improvements, respectively
\item Natural language consultation enhances collaboration; context engineering reduces hallucinations
\item Platform provides an intuitive interface for data exploration and visualization
\end{highlights}


\maketitle

\section{Introduction}

Table Question Answering (TableQA) enables users to pose natural language questions about structured tabular data for extracting relevant information and insights~\citep{SurveyTableQA2023}. Unlike traditional text-based Question Answering (QA), TableQA requires an understanding of table structure, including rows, columns, cell contents, and semantic relationships, for accurate answer retrieval. By leveraging tabular resources, this task facilitates natural language querying of complex structured information with practical applicability in various scenarios.

Early studies in TableQA primarily adopted rule-based and retrieval-based methods~\citep{RuleBasedQA2021, TableExtraction2019}, followed by deep learning (DL) approaches such as TAPAS, TAPEX and TaBERT that jointly encode natural language and table schemas~\citep{TAPAS2020, TaBERT2020}. While these methods achieved improvements, they continue to face challenges in interpretability and complex reasoning when handling multi-hop queries and semantic relationship inference.

Recent advances have introduced large language models (LLMs) to TableQA through direct prompting and code generation paradigms~\citep{LLMTableProcessing2023, LLMTabularSurvey2023}. However, existing approaches face critical challenges including context length limitations, hallucination issues, and difficulties in complex reasoning scenarios involving semantic relationships and multi-step logic~\citep{nan-etal-2022-fetaqa, TableMeetsLLM2024}.

To address these issues, we propose an LLM-based multi-agent DataFactory framework for TableQA that: (1) automates table ingestion with LLM-assisted schema understanding; (2) formalizes an automated data-to-knowledge-graph transformation via the mapping $\mathcal{T}: \mathcal{D} \times \mathcal{S} \times \mathcal{R} \rightarrow \mathcal{G}$ for capturing semantic relationships; (3) reduces hallucinations through context engineering and few-shot prompting that integrates historical QA, Data Definition Language (DDL)/graph schemas, and domain knowledge; and (4) enables multi-step reasoning by jointly leveraging structured (Structured Query Language (SQL)) and relational (Cypher) retrieval within a fully autonomous pipeline.

The proposed DataFactory framework comprises a Data Leader and two specialist teams (Database and Knowledge Graph) and operates in a three-phase workflow. In the information storage phase, information processing agents analyze headers, field names, and types to ingest data into databases and automatically construct knowledge graphs from tables using our rules and algorithms. In the knowledge extraction phase, retrieval agents employ historical QA records, DDL/graph schemas, and domain knowledge to craft context-engineered prompts and generate SQL/Cypher queries. Their outputs are then interpreted by analysis agents and, when appropriate, visualized as statistical charts or subgraphs. In the insight generation phase, the Data Leader applies the ReAct paradigm to decompose user questions into subtasks (data exploration, strategy formulation, answer synthesis), dispatches them to teams, and integrates results into coherent responses.

\begin{figure}
\centering
\includegraphics[width=1.0\textwidth]{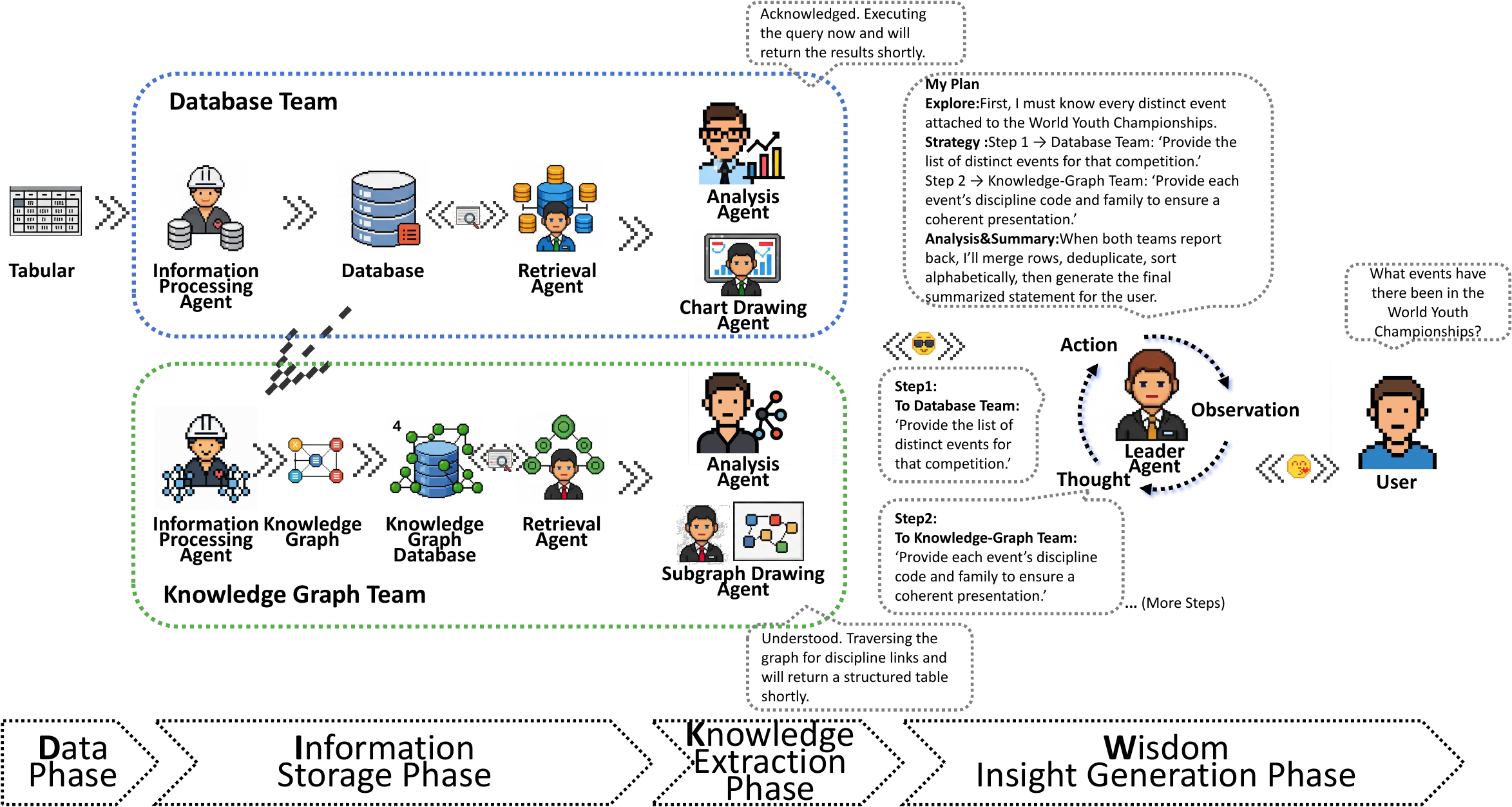}
\caption{Overview of the LLM-based multi-agent DataFactory framework for TableQA. The system consists of a Database Team, a Knowledge Graph Team, and a Data Leader that orchestrates their collaboration from an initial data phase through three processing phases: information storage, knowledge extraction, and insight generation.}
\label{FIG:datafactory}
\end{figure}

In summary, our contributions are: (1) specialized team coordination mechanisms that establish dedicated Database and Knowledge Graph teams for systematic decomposition of TableQA tasks into complementary reasoning modalities, moving beyond single-agent limitations through natural language-based consultation and adaptive strategy adjustment; (2) automated knowledge integration through formalized data-to-knowledge graph transformation $\mathcal{T}: \mathcal{D} \times \mathcal{S} \times \mathcal{R} \rightarrow \mathcal{G}$ that enables consistent entity resolution, semantic relationship discovery, and scalable system architecture for real-world deployment scenarios; and (3) dynamic reasoning orchestration via a Data Leader employing the ReAct paradigm to coordinate between structured retrieval and relational inference, under which we conduct systematic experiments that jointly evaluate performance and scalability through comparisons with representative baselines, cross-provider and cross-size model settings, collaboration pattern analysis, and ablation studies. The framework is further supported by an interactive platform\footnote{Platform demonstration website: \url{https://wisdomindata.netlify.app/}. This website presents interactive demonstration videos and will host open-source code upon completion of integration.} that provides comprehensive interfaces for data exploration, knowledge graph visualization, and interactive multi-agent collaboration, allowing users to engage separately with the Database and Knowledge Graph teams or orchestrate joint workflow via the Data Leader.

\section{Related Work}

\subsection{LLM-based TableQA Methods}

Currently, LLMs are applied to TableQA through two main paradigms: direct prompting methods and code generation methods~\citep{LLMTableProcessing2023, LLMTabularSurvey2023}. 

\textbf{Direct Prompting Methods.} Direct prompting involves providing table content (often suitably formatted) as context alongside the question to an LLM, allowing the model to directly generate the answer. However, this approach faces significant challenges: (1) context length limitations that restrict data handling capabilities~\citep{TabSD2023,PieTa2024}; (2) hallucination issues where models generate content not supported by the table data~\citep{HallucinationDetection2024,TIDE2025,sairaj2025ontology}; and (3) difficulty in precise data localization and low interpretability of the end-to-end reasoning process, which hinders the explicit explanation of answer sources or derivations~\citep{InterpretableLLMTQA2023,CTP2023,E5TableAnalysis2025}. Recent work by \citet{sairaj2025ensemble} and \citet{sairaj2025ontology} demonstrates that LLMs tend to capture lexical co-occurrence patterns rather than genuine entity relationships, leading to factual hallucinations and topic redundancy in generated content. Some approaches attempt fine-tuning to improve accuracy but show limited success in addressing these core challenges while also increasing practical complexity~\citep{liu2022tapex,zhang-etal-2024-tablellama,yang2025tabler1inferencetimescalingtable}.

\textbf{Code Generation Methods.} The second paradigm involves LLMs generating executable program code, converting the QA task into executable operations that manipulate tables through external tools, thereby providing transparency and verifiability. This paradigm includes: (1) \textit{Python code generation}, which utilizes libraries such as pandas for filtering and computation~\citep{Lang2CodeQA2025,ReAcTable2024}, though this approach depends on runtime environments and exhibits performance degradation with hierarchical table structures or sparse data distributions~\citep{APIAssistedCodeGen2023}; (2) \textit{Spreadsheet formula generation}, where LLMs generate formulas requiring accurate cell referencing and formula syntax adherence—execution typically requires interaction with Excel, which complicates implementation~\citep{AnswerFormulaGeneration2023}; and (3) \textit{SQL query generation}, which translates natural language into SQL queries but requires database schema design and presents limitations with multi-step reasoning tasks~\citep{InterpretableLLMTQA2023,OpenQATablesText2019, TextToSQLSurvey2022,nahid-rafiei-2024-table,GUO2025103978}.

Overall, both paradigms emphasize structured field extraction, but they remain limited in addressing complex reasoning that involves semantic relationships, cross-row synthesis, or multi-row numerical comparisons~\citep{nan-etal-2022-fetaqa,TableMeetsLLM2024}. These limitations have motivated the exploration of agent-based frameworks that introduce explicit reasoning and action mechanisms.  

\subsection{Agent-based TableQA Methods}

Agent-based approaches decompose complex reasoning into iterative cycles of reasoning and acting, thereby improving interpretability and reducing errors compared to single-pass LLM methods. Existing work has explored several technical directions:  

\textbf{ReAct-style decomposition.}  
Inspired by the \textsc{ReAct} paradigm~\citep{ReAct2023}, these frameworks interleave reasoning steps and tool invocations to navigate tables systematically~\citep{ReAcTable2024}. A typical workflow includes identifying relevant rows or columns, executing SQL/Python queries, reflecting on intermediate outputs, and refining until the answer is confirmed. Key challenges remain in managing multi-hop reasoning complexity and ensuring reliable reflection mechanisms.  

\textbf{Single-agent refinement.} Various single-agent frameworks have been proposed to improve reasoning quality through iterative refinement and self-correction mechanisms. These approaches typically employ role-based reasoning~\citep{TQAgent2025} or multi-step verification to enhance answer accuracy~\citep{TableCritic2025}. While effective in reducing certain types of errors, these methods face limitations in handling complex queries that require diverse expertise and parallel processing capabilities.

\textbf{Knowledge augmentation.}  
Some methods enhance reasoning by integrating external knowledge sources or graph-based representations~\citep{TQAgent2025,GraphOTTER2025,MEIER2025104202}. However, these approaches typically emphasize partial table content or shallow inter-table links, offering only limited coverage of semantic relationships. Their integration of external knowledge often remains superficial, making deep multi-hop reasoning difficult.  

\textbf{Tool orchestration and planning.}  
Planning-based systems selectively invoke heterogeneous tools, such as SQL for retrieval and LLMs for reasoning~\citep{Weaver2025,nahid-rafiei-2024-table,OJURI2025104136}. This improves transparency and efficiency, though challenges persist in maintaining global coherence and recovering from execution failures. Recent frameworks such as \textbf{StructGPT} design interfaces for tables, knowledge graphs, and databases~\citep{StructGPT2023}. Yet, these interfaces remain tied to separate data sources rather than forming an integrated workflow for unified TableQA.  

\textbf{Multi-agent collaboration.}  
More recent studies employ multi-agent pipelines with explicit role specialization. For example, \textbf{AutoPrep} decomposes data preparation into Planner, Programmer, and Executor roles~\citep{AutoPrep2024}. \textbf{AutoTQA} further extends this idea by coordinating multiple LLM agents for end-to-end TableQA~\citep{AutoTQA2024}. Similarly, \textbf{MACT} demonstrates that combining planning and coding agents with external tools can achieve competitive accuracy without fine-tuning~\citep{zhou-etal-2025-efficient}. 

However, recent empirical analysis by \citet{xie2024multiagentwhy} reveals that multi-agent systems often fail to outperform single-agent baselines, with failures primarily attributed to coordination problems, specification violations, and inadequate verification mechanisms rather than individual LLM limitations. Their Multi-Agent System Failure Taxonomy (MAST) identifies 14 distinct failure modes across specification design, inter-agent misalignment, and task validation, suggesting that current approaches require fundamental architectural improvements rather than tactical prompt engineering fixes. 

Despite these challenges, existing systems highlight the promise of division of labor, though their focus remains largely on table manipulation and database querying, with coordination primarily based on predetermined workflow execution rather than dynamic consultation and reasoning. Our \textbf{DataFactory} framework advances this research direction by introducing a dedicated Knowledge Graph team, enabling data-to-KG transformation and integrated reasoning across SQL and graph queries. More importantly, our approach enables natural language-based consultation and deliberation among three specialized agents—Data Leader, Database Team, and Knowledge Graph Team—rather than rigid workflow-based task distribution, allowing for more flexible and adaptive problem-solving strategies.  

From a scenario perspective, fixed procedural multi-agent frameworks are preferable for well-specified pipelines with predictable steps and strict resource budgets, offering deterministic execution and high stability. In contrast, natural language-based collaboration—as adopted in our framework—benefits complex or exploratory queries that demand adaptive planning, multi-hop reasoning across SQL and graph queries, and dynamic strategy adjustment based on intermediate evidence. These paradigms are complementary; our research targets the latter setting.

In summary, despite the promising capabilities of agent-based TableQA methods, several critical challenges persist that motivate our multi-agent data factory design: 

(1) \textbf{Lack of specialized team coordination}: Existing approaches primarily rely on single-agent systems with limited role specialization, which constrains the exploitation of complementary strengths between structured data processing and relational knowledge representation. 

(2) \textbf{Insufficient knowledge integration}: Current knowledge augmentation methods focus on partial table content or superficial inter-table relationships, lacking comprehensive understanding of data relationships and deep semantic connections between entities. 

(3) \textbf{Inadequate reasoning orchestration}: Current frameworks struggle to dynamically coordinate between different data access modalities, missing opportunities for sophisticated multi-step reasoning that combines both structured retrieval and relational inference. 

(4) \textbf{Limited natural language consultation mechanisms}: Existing multi-agent systems primarily employ rigid workflow-based coordination, lacking flexible natural language consultation and deliberation capabilities that enable adaptive strategy adjustment and cross-team knowledge sharing during complex reasoning processes.

(5) \textbf{Limited automation and scalability}: Few frameworks support systematic data ingestion or knowledge graph construction, constraining real-world deployment.

\section{Research Objectives}

Building on the above analysis, this paper aims to develop a comprehensive solution that targets these limitations while advancing multi-agent reasoning for TableQA. Specifically, our research targets the fundamental issues of specialization, coordination, and scalability through the following objectives:

(1) \textbf{Establish Specialized Team Coordination Mechanisms}: We propose dedicated Database and Knowledge Graph teams that leverage the complementary strengths of structured data processing and relational knowledge representation, moving beyond single-agent limitations to achieve systematic task decomposition. This approach aims to mitigate the coordination failures highlighted in MAST by implementing clear role boundaries and natural language consultation protocols.

(2) \textbf{Formalize Automated Knowledge Integration}: We aim to create robust cross-source alignment through automated data-to-knowledge graph transformation algorithms that ensure consistent entity resolution and semantic integration, thereby enabling comprehensive data understanding.  

(3) \textbf{Design Sophisticated Reasoning Orchestration}: We introduce a Data Leader that employs the ReAct paradigm to dynamically coordinate between structured retrieval and relational inference, thus enabling complex multi-hop reasoning.  

(4) \textbf{Implement Natural Language-Based Consultation Mechanisms}: We develop a flexible inter-agent communication mechanism that enables natural language consultation and deliberation among specialized teams, facilitating adaptive strategy adjustment and knowledge sharing during complex reasoning processes rather than rigid workflow execution.

(5) \textbf{Achieve Enhanced System Scalability}: We aim to develop fully autonomous data ingestion and knowledge graph construction pipelines that can address complex TableQA scenarios involving multi-hop reasoning, semantic relationship analysis, and knowledge integration in real-world deployments.

\section{Methodology}

This section presents a comprehensive exposition of the internal mechanisms and core algorithms underlying our proposed multi-agent data factory. We begin with an overview of the system's collaborative architecture in Section~\ref{sec:overview}, providing a high-level blueprint of each core component. The subsequent sections examine the specific implementation of each component: Section~\ref{sec:database-team} details how the Database Team processes structured data and retrieves targeted information with corresponding insights; Section~\ref{sec:knowledge-graph-team} describes how the Knowledge Graph Team performs relational knowledge representation and reasoning; and Section~\ref{sec:data-leader} explains how the Data Leader decomposes problem-solving processes and coordinates both teams to generate comprehensive insights.

\subsection{Overview}\label{sec:overview}

DataFactory introduces a tripartite collaborative architecture that decomposes complex TableQA tasks through specialized expertise and intelligent coordination. As illustrated in Figure~\ref{FIG:datafactory}, the framework addresses our research objectives through three coordinated components: a Data Leader that orchestrates natural language-based consultation mechanisms, a Database Team specializing in structured data processing, and a Knowledge Graph Team focusing on relational knowledge representation.

The Data Leader implements sophisticated reasoning orchestration by employing the ReAct paradigm to decompose user queries into structured and relational reasoning subtasks. Rather than following predetermined workflows, the Data Leader engages in dynamic consultation with specialist teams, enabling adaptive strategy adjustment based on query complexity and intermediate findings. This natural language-based coordination mechanism represents a key advancement over rigid task distribution approaches in existing multi-agent systems.

The Database Team establishes specialized coordination for structured data operations through context-enhanced SQL generation that integrates historical QA patterns, schema information, and domain knowledge. The team's core innovation lies in automated data ingestion combined with retrieval-augmented generation techniques that reduce hallucination while maintaining query accuracy for numerical computations and aggregations.

The Knowledge Graph Team enables automated knowledge integration through our formalized data-to-knowledge graph transformation $\mathcal{T}: \mathcal{D} \times \mathcal{S} \times \mathcal{R} \rightarrow \mathcal{G}$, where tabular data $\mathcal{D}$, schema definitions $\mathcal{S}$, and relationship patterns $\mathcal{R}$ are systematically transformed into knowledge graph $\mathcal{G}$. This automated transformation addresses the scalability limitation by enabling systematic semantic relationship discovery and multi-hop reasoning capabilities that extend beyond conventional structured querying.

The framework operates through three coordinated phases that transform raw data into actionable insights. During the information storage phase, automated data ingestion establishes both structured databases and semantic knowledge graphs through LLM-assisted schema understanding and relationship discovery. The knowledge extraction phase employs context-enhanced prompting strategies that integrate historical patterns, schema information, and domain knowledge to generate accurate SQL and Cypher queries while reducing hallucination risks. The insight generation phase coordinates multi-agent collaboration through ReAct-based decomposition, where the Data Leader dynamically assigns reasoning subtasks and synthesizes results from both structural and relational perspectives.

This architectural design addresses the fundamental challenge of combining complementary reasoning modalities within a unified framework. By decoupling TableQA into structured querying and relational reasoning capabilities, the framework enables sophisticated multi-hop reasoning that leverages the strengths of both paradigms while maintaining interpretability through explicit coordination mechanisms.

\subsection{Database Team}\label{sec:database-team}

The Database Team establishes specialized coordination mechanisms for structured data operations, addressing the first research objective through dedicated expertise in numerical computation, aggregation, and precise data filtering. The team's architectural innovation lies in integrating LLM-assisted semantic understanding with traditional database operations, enabling automated data ingestion while maintaining query accuracy through context-enhanced prompting strategies.

The team comprises four specialized agents that collectively implement our approach to reducing hallucination risks while preserving computational efficiency. The Information Processing Agent enables automated data ingestion through hybrid LLM-rule integration, the Information Retrieval Agent implements context-enhanced Text-to-SQL generation using retrieval-augmented prompting, the Information Analysis Agent provides semantic interpretation of structured results, and the Visualization Agent supports analytical insight presentation. This specialization directly addresses the limitation of single-agent approaches in handling diverse structured data processing requirements.

\subsubsection{Database Information Processing Agent}\label{sec:db-info-processing-agent}

The Database Information Processing Agent implements automated knowledge integration for structured data through a hybrid approach that combines LLM-based semantic understanding with rule-based optimization. This design addresses the scalability limitation identified in our research objectives by enabling systematic data ingestion without manual schema engineering while maintaining data integrity and query performance.

The agent's core innovation lies in intelligent schema understanding that leverages LLMs for semantic parsing of heterogeneous table formats while employing statistical analysis and rule-based validation for robustness. Unlike conventional data processing pipelines that rely solely on heuristic rules, our approach enables adaptive handling of inconsistent header formats and ambiguous data types through semantic context analysis. The LLM component analyzes table structure and content semantics to identify entity relationships and data dependencies, while rule-based components ensure syntactic correctness and constraint satisfaction.

The automated DDL generation process represents a key advancement in bridging semantic understanding with database schema design. The agent analyzes observed data characteristics to recommend appropriate constraints, indexes, and relationships, reducing manual schema engineering effort while maintaining query optimization properties.

\begin{figure}
	\centering
		\includegraphics[width=1.0\textwidth]{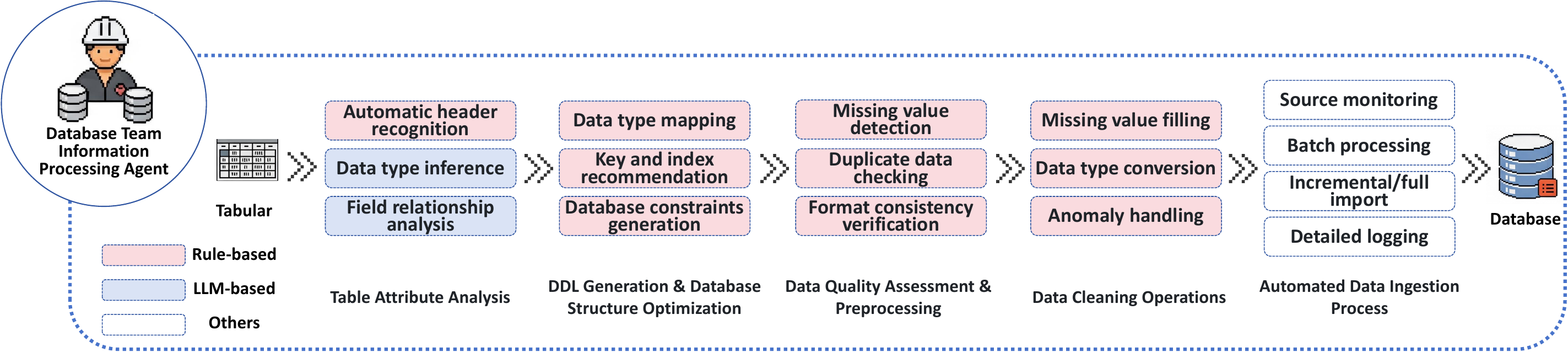}
	\caption{Workflow of the Database Information Processing Agent. The automated pipeline combines rule-based operations for schema construction with LLM-based analysis for semantic understanding, covering table analysis, DDL generation, data ingestion, and quality assessment.}
	\label{FIG:db-info-processing}
\end{figure}

Quality assurance mechanisms integrate LLM-based judgments with preset validation rules to handle data inconsistencies and missing value patterns. The agent adapts cleaning strategies based on data characteristics and domain requirements, providing flexibility that static rule-based approaches cannot achieve. This adaptive quality control contributes to ensuring data consistency and completeness (see Figure~\ref{FIG:db-info-processing}).

\subsubsection{Database Information Retrieval Agent}\label{sec:db-info-retrieval-agent}

The Database Information Retrieval Agent implements context-enhanced Text-to-SQL generation to address hallucination risks through multi-source knowledge integration. This agent represents a key advancement in sophisticated reasoning orchestration by systematically incorporating historical patterns, schema information, and domain knowledge into query generation processes.

The agent's core innovation lies in adaptive context construction that extends traditional Text-to-SQL approaches through retrieval-augmented prompting strategies. Rather than relying on static examples or schema-only guidance, the agent dynamically assembles contextual information from three complementary sources. Historical QA retrieval leverages semantic similarity to identify relevant query patterns from validated question-SQL pairs, enabling the system to learn from successful reasoning precedents while avoiding repetition of previous errors. To continuously build this historical knowledge base, the system automatically saves each user query along with its corresponding generated SQL and execution results after every interaction, storing them as vectorized representations in a dedicated vector database for efficient semantic retrieval. Schema-aware integration ensures generated queries align with actual database structures through comprehensive DDL incorporation, while domain knowledge injection enables sophisticated interpretation of business terminology and calculation conventions.

The retrieval-augmented generation (RAG) mechanism represents a significant methodological contribution to reducing hallucination in SQL generation. Unlike conventional approaches that select examples randomly or use predetermined templates, our agent employs semantic embedding and structural compatibility scoring to identify contextually relevant precedents. This adaptive selection process filters examples based on semantic relatedness and syntactic pattern compatibility, ensuring that few-shot prompts provide both meaningful guidance and accurate structural templates for the current query generation task.

\begin{figure}
	\centering
		\includegraphics[width=1.0\textwidth]{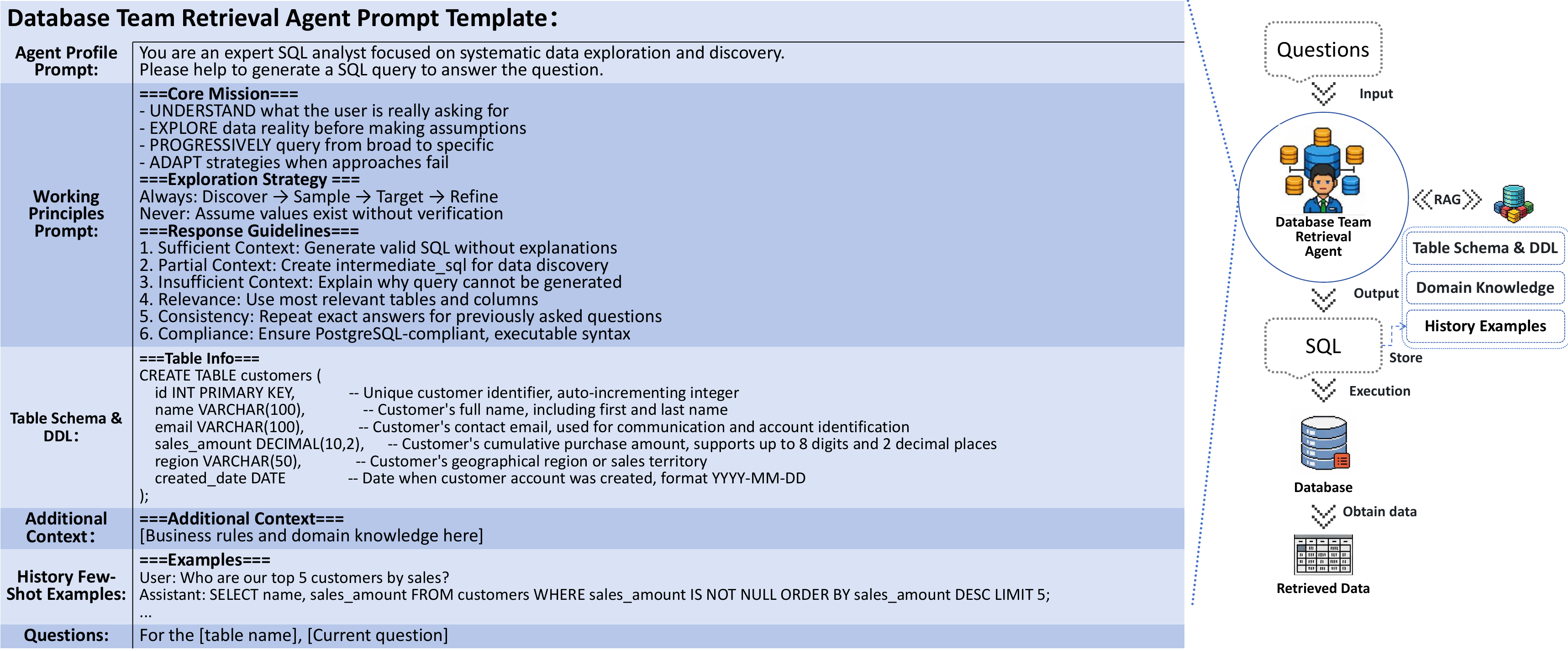}
	\caption{Architecture of the Database Information Retrieval Agent. The agent integrates table schema and DDL information, domain knowledge, and retrieved historical question-SQL pairs to construct prompts for SQL generation and database querying.}
	\label{FIG:db-retrieval-agent}
\end{figure}

Multi-source information fusion enables the agent to generate SQL queries that satisfy multiple correctness criteria simultaneously. The synthesis process ensures semantic consistency with user intent, syntactic correctness for the target database schema, and logical coherence with domain-specific business requirements (see Figure~\ref{FIG:db-retrieval-agent}).

\subsubsection{Database Information Analysis Agent and Visualization Agent}\label{sec:db-info-analysis-agent}

The Database Information Analysis Agent interprets SQL query results and generates natural language explanations for end users. This agent addresses the limitation of traditional database systems that return raw tabular data without contextual interpretation, enabling semantic understanding of query outcomes within the broader analytical context.

The agent processes structured query results through systematic data interpretation that identifies key metrics, dimensions, and statistical patterns. For analytical queries involving aggregation, grouping, or sorting operations, the agent analyzes the business significance of these computations and integrates them with the original user query context. This interpretation process extends beyond simple data presentation to provide meaningful analytical insights.

Natural language response generation combines query results with user intent to produce coherent explanations that summarize findings and provide relevant context. Rather than returning isolated values, the agent constructs responses that include supporting evidence and comparative information when appropriate. For numerical results, the agent performs basic statistical computations and identifies notable trends or anomalies that enhance user understanding.

The agent handles cross-row analysis requirements by integrating information from multiple result rows to derive comprehensive conclusions. This capability supports complex business intelligence queries that require synthesis of patterns across different data dimensions, enabling users to understand relationships and trends that emerge from the data.

\begin{figure}
	\centering
	\begin{subfigure}[b]{0.75\textwidth}
		\centering
		\includegraphics[width=\textwidth]{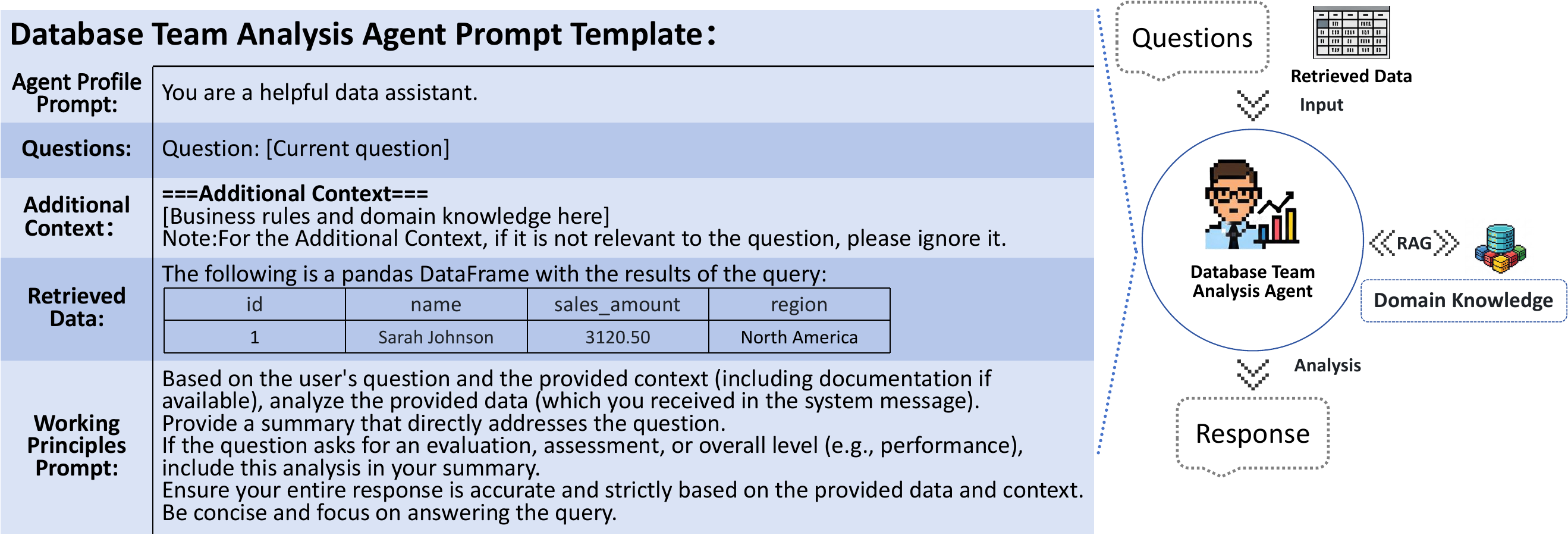}
		\caption{Database Information Analysis Agent with prompt template}
		\label{FIG:db-analysis-agent}
	\end{subfigure}
	
	\begin{subfigure}[b]{0.75\textwidth}
		\centering
		\includegraphics[width=\textwidth]{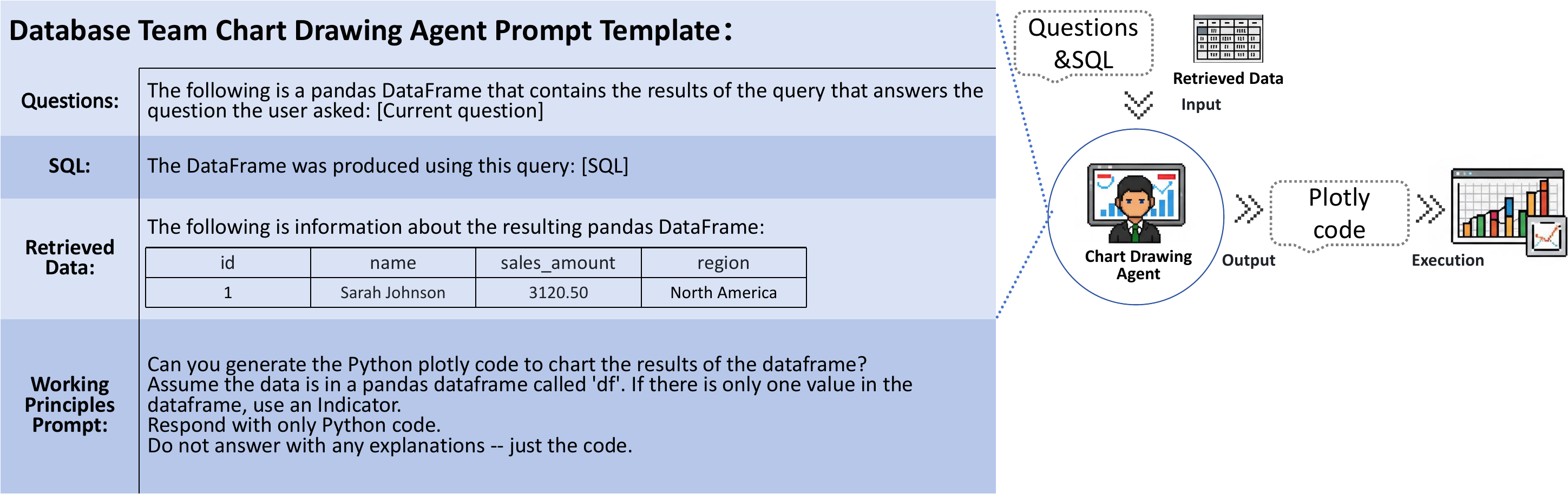}
		\caption{Database Visualization Agent with prompt template}
		\label{FIG:db-visualization-agent}
	\end{subfigure}
	\caption{Database Information Analysis and Visualization Agents. (a) The Analysis Agent transforms retrieved tables into natural-language summaries with domain context. (b) The Visualization Agent generates plotting code to produce interactive charts for data exploration.}
	\label{FIG:db-analysis-visualization}
\end{figure}

The Visualization Agent complements the analysis process by generating appropriate charts when visual representation enhances data comprehension. The agent evaluates query results and user requirements to determine optimal visualization approaches, then generates plotting code for interactive display through the web interface (see Figure~\ref{FIG:db-analysis-visualization}).

\subsection{Knowledge Graph Team}\label{sec:knowledge-graph-team}

The Knowledge Graph Team implements automated knowledge integration through relational data representation, addressing the second research objective by transforming flat tabular data into semantic networks that support multi-hop reasoning and relationship discovery. This team specializes in graph-based query processing when user questions require an understanding of entity relationships and indirect associations between data elements.

The team comprises four specialized agents that handle distinct aspects of graph-based reasoning: Information Processing agents construct knowledge graphs from tabular data, Information Retrieval agents generate Cypher queries for graph traversal, Information Analysis agents interpret graph query results, and Visualization agents create visual representations of the knowledge graphs. This division enables systematic handling of relationship-centered queries that extend beyond conventional structured data aggregation.

\subsubsection{Knowledge Graph Information Processing Agent}\label{sec:kg-info-processing-agent}

The Knowledge Graph Information Processing Agent enables automated knowledge integration through systematic transformation of tabular data into semantic knowledge graphs. This agent addresses a fundamental limitation in existing TableQA approaches: the inability to capture and utilize semantic relationships between entities for complex multi-hop reasoning.

Our key innovation lies in formalizing automated data-to-knowledge graph transformation via the mapping function:
\begin{equation}
\mathcal{T}: \mathcal{D} \times \mathcal{S} \times \mathcal{R} \rightarrow \mathcal{G}
\end{equation}
where $\mathcal{D} = \{d_1, d_2, \ldots, d_n\}$ represents the input tabular dataset, $\mathcal{S}$ denotes the entity schema patterns that specify how to identify and extract entities from table rows, $\mathcal{R}$ captures the relationship rules that determine how entities connect semantically, and $\mathcal{G} = (V, E, A)$ represents the resulting knowledge graph with nodes $V$, edges $E$, and attributes $A$. Unlike existing approaches that rely on manual schema engineering or heuristic rules, our transformation enables systematic semantic relationship discovery through LLM-assisted pattern recognition combined with algorithmic validation.

This process follows the ``identify-construct-connect-optimize'' paradigm, ensuring both efficiency and accuracy. Conceptually, it consists of two stages: \textit{entity construction} and \textit{relationship discovery}. In practice, implementation proceeds through three workflow stages: strategy generation, configuration validation, and graph construction.

\textbf{(1) Entity Construction Theoretical Framework}

\textbf{Entity Identifier Generation Mechanism.} To ensure entity uniqueness and traceability, we define an identifier (ID) generation function:
\begin{equation}
\text{ID}(row, type) = \begin{cases}
type \oplus \text{Extract}(row), & \text{basic mode} \\
namespace \oplus type \oplus \text{Extract}(row), & \text{with namespace}
\end{cases}
\end{equation}
where $\text{Extract}(row)$ extracts identifier values from the data row, and $\oplus$ denotes string concatenation with appropriate separators. For single identifier columns, $\text{Extract}(row)$ returns the column value directly. For multiple identifier columns, it concatenates them with colons to form a composite identifier.

\textbf{Split-based Entity Construction Strategy.} For processing composite cell data containing multiple values, we define:
\begin{equation}
\text{Entities}(row, column) = \begin{cases}
\{\text{Split}(row[column], \delta)\}, & \text{if delimiters detected} \\
\{row[column]\}, & \text{otherwise}
\end{cases}
\end{equation}
where $\text{Split}(value, \delta)$ splits a cell value by delimiter set $\delta = \{\text{``,''},\text{``;''},\text{``|''},\ldots\}$ into individual entity instances. For example, a cell containing ``AI,Machine Learning,Data Science'' generates three separate technology entities.

\textbf{Entity Attribute Construction.} For each entity, we aggregate attributes from multiple sources:
\begin{equation}
\text{Attributes}(entity) = A_{core} \cup A_{custom} \cup A_{meta}
\end{equation}
where:
\begin{itemize}
    \item $A_{core}$ contains the primary identifier attributes for entity traceability
    \item $A_{custom}$ includes user-specified attribute columns mapped to the entity
    \item $A_{meta}$ stores system-generated metadata (timestamps, source information, etc.)
\end{itemize}

\textbf{Entity Merging Strategy.} When duplicate entities are detected, we apply an intelligent merging function:
\begin{equation}
\text{Merge}(e_1, e_2) = (id_1, \text{Combine}(A_1, A_2), R_1 \cup R_2)
\end{equation}
where $\text{Combine}(A_1, A_2)$ implements a priority-based attribute merging strategy: non-null values override null values, and for conflicts between non-null values, the first occurrence is retained with conflict logging.

\textbf{(2) Relationship Discovery and Establishment Theoretical Framework}

\textbf{Intra-row Relationship Matching Mechanism.} Intra-row relationship matching is the most intuitive relationship establishment pattern, creating semantic connections between different entities within a single data row. The working principle of this mechanism is based on the assumption of ``co-occurrence implies correlation'': different types of entities appearing in the same data row have some semantic association.

For data row $d_i$, we define the intra-row relationship generation as:
\begin{equation}
R_{\text{same}}(d_i) = \{(e_s, e_t, r) \mid e_s, e_t \in \text{Entities}(d_i), \text{type}(e_s) \neq \text{type}(e_t), r \in \mathcal{R}_{\text{intra}}\}
\end{equation}
where $\text{Entities}(d_i)$ returns all entities extracted from row $d_i$, and $\mathcal{R}_{\text{intra}}$ defines the allowed intra-row relationship types between different entity types.

\textbf{Cross-row Relationship Matching and Grouping Strategy.} Cross-row relationship matching enables the discovery of complex semantic connections beyond single-row data scope, including hierarchical relationships, dependency relationships, and temporal relationships. The system adopts a grouping-based cross-row matching strategy, defined as follows:

For cross-row relationship discovery, we first partition the dataset into logical groups:
\begin{equation}
\text{Group}_k = \{d_i \mid d_i[\text{col}] = \text{value}_k, \forall \text{col} \in C_{\text{group}}\}
\end{equation}
where $C_{\text{group}}$ represents the grouping column set and $\text{value}_k$ is the shared value for group $k$. This grouping strategy is based on the principle that entities sharing similar contextual attributes are more likely to have meaningful relationships.

\textbf{Multi-pattern Rule Evaluation Engine.} The establishment of cross-row relationships relies on sophisticated rule evaluation mechanisms, with the system supporting two rule types:
For semantic similarity-based relationships, we define:
\begin{equation}
\text{Similarity}(e_1, e_2) = \frac{\text{embed}(e_1) \cdot \text{embed}(e_2)}{||\text{embed}(e_1)|| \cdot ||\text{embed}(e_2)||}
\end{equation}
A relationship is established when $\text{Similarity}(e_1, e_2) \geq \theta$, where $\theta$ is a predefined threshold. 

For attribute-based comparisons, we use standard comparison operators ($=$, $>$, $<$, $\geq$, $\leq$) between corresponding entity attributes.

\textbf{Composite Logical Expression Processing.} In practical applications, relationship establishment often requires satisfying complex combinations of logical conditions. The system supports the construction of sophisticated conditional expressions through logical operators (AND, OR) and grouping mechanisms. The operational mechanism adopts a recursive evaluation strategy: for rule sets containing multiple conditions, the system performs recursive evaluation according to the priority of logical operators and grouping structure, supporting arbitrary depth of logical nesting.

The transformation process requires structured input specification comprising tabular data $\mathcal{D}\in\mathbb{R}^{n\times m}$, entity schema patterns $\mathcal{S}=\{(\tau_i, C_{\text{id}}^{(i)}, C_{\text{attr}}^{(i)}, \text{split\_config}^{(i)})\}$ that define entity extraction rules, and relationship rules $\mathcal{R}=\{(r_j,\tau_{\text{src}}^{(j)},\tau_{\text{tgt}}^{(j)},\text{match\_mode}^{(j)},\text{rules}^{(j)})\}$ that specify semantic connection patterns. The resulting knowledge graph $\mathcal{G} = (V, E, A)$ provides structured semantic representation supporting complex query operations and multi-hop reasoning.

\begin{figure}
	\centering
	\includegraphics[width=1.0\textwidth]{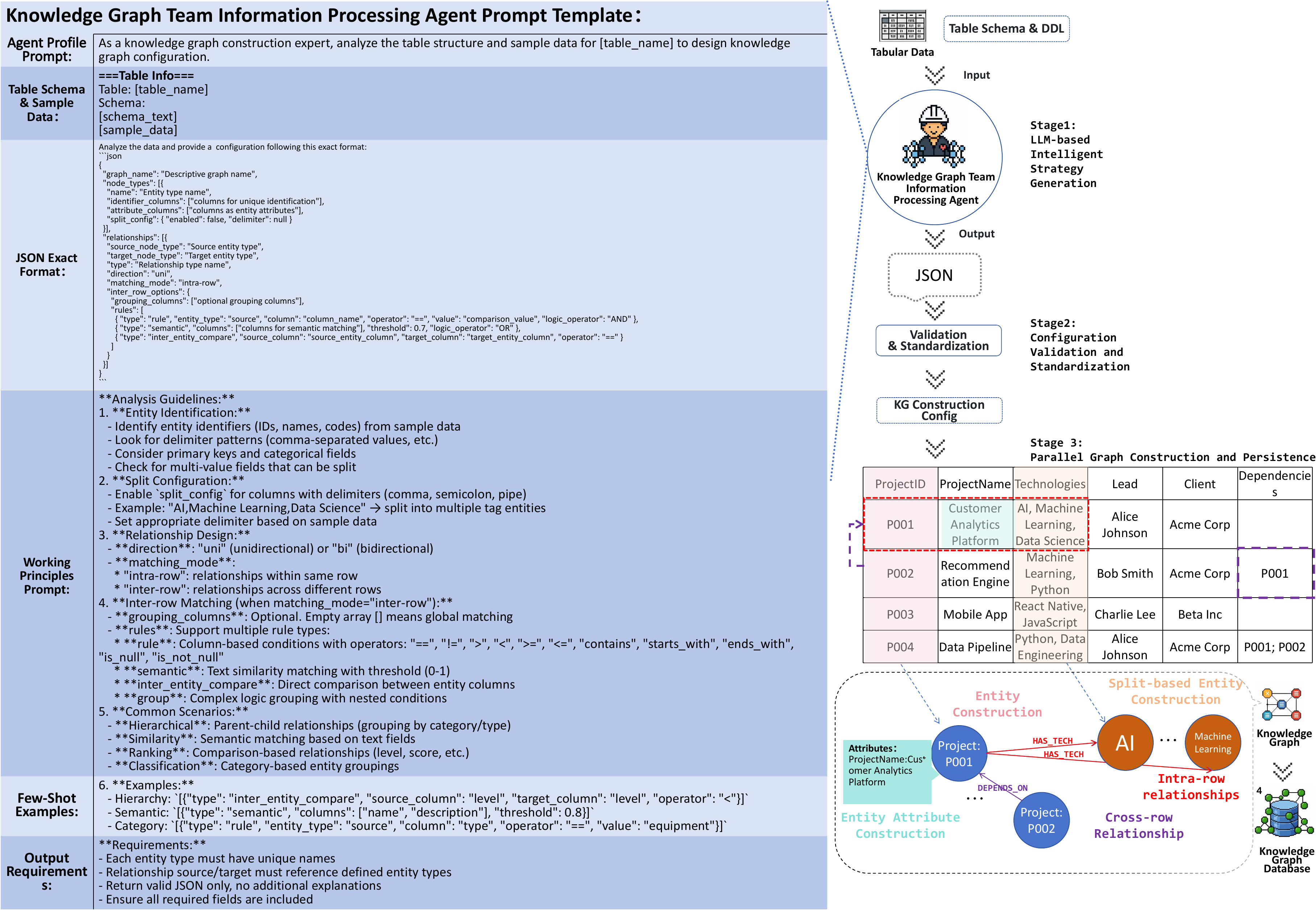}
	\caption{Workflow of the Knowledge Graph Information Processing Agent. The agent follows a three-stage process of strategy generation, configuration validation, and graph construction to transform tabular data into a knowledge graph.}
	\label{FIG:kg-info-processing}
\end{figure}

Implementation follows a three-stage workflow (Figure~\ref{FIG:kg-info-processing}) that balances automation with quality assurance. The strategy generation stage employs LLM-based analysis of table schemas and sample data to determine optimal entity schema patterns $\mathcal{S}$ and relationship rules $\mathcal{R}$. This avoids manual configuration while maintaining semantic accuracy. Configuration validation ensures algorithmic consistency and parameter validity. Graph construction applies the entity identification and relationship discovery algorithms in parallel, exporting results to Neo4j~\citep{noauthororeditorneo4j} through optimized batch operations.

This design achieves automated knowledge integration that scales beyond manual schema engineering limitations while maintaining the semantic richness required for complex TableQA scenarios. The approach enables systematic discovery of entity relationships that traditional structured query approaches cannot capture, supporting the multi-hop reasoning capabilities essential for advanced question answering tasks.

The strategy generation stage performs structure-aware analysis using table schemas and sample data to produce semantic understanding through LLM analysis. Unlike heuristic approaches, this stage generates precise entity extraction and relationship discovery configurations based on actual data characteristics. Configuration validation ensures algorithmic consistency through systematic verification of entity definitions and relationship rules. The graph construction stage implements the formalized algorithms for entity identification and relationship establishment, utilizing in-memory computation with batch persistence to Neo4j graph database. Error handling mechanisms in the graph construction stage provide automatic fallback and state recovery, ensuring robust operation across diverse table formats and data quality scenarios.

\subsubsection{Knowledge Graph Information Retrieval Agent}\label{sec:kg-info-retrieval-agent}

The Knowledge Graph Information Retrieval Agent implements context-enhanced Text-to-Cypher generation that extends structured query capabilities to relational reasoning tasks. This agent addresses the limitation of traditional approaches in handling multi-hop queries and semantic relationships by enabling natural language interaction with graph structures.

The agent's core innovation lies in adaptive context construction for graph query generation. Unlike static schema approaches, our method dynamically assembles contextual information from graph metadata, historical query patterns, and domain knowledge. Dynamic graph schema extraction queries Neo4j metadata interfaces to obtain current node labels, relationship types, and property keys, providing the LLM with a comprehensive understanding of available graph structures and traversal possibilities.

Context-enhanced prompting integrates three components that collectively improve query accuracy. Historical query retrieval employs semantic similarity matching to identify relevant question-Cypher pairs from validated examples, filtering results based on structural compatibility including node label overlap and relationship patterns. The system maintains a continuously growing repository of historical QA data by automatically preserving each user query, its generated Cypher statement, and execution results after every interaction, with all entries stored as vectorized embeddings in a vector database to enable efficient semantic similarity-based retrieval for future query enhancement. Domain knowledge integration provides semantic context that enables the LLM to understand entity meanings and relationship semantics within specific application domains. External knowledge injection helps generate queries that incorporate domain-specific reasoning patterns beyond basic graph traversal.

\begin{figure}
	\centering
	\includegraphics[width=1.0\textwidth]{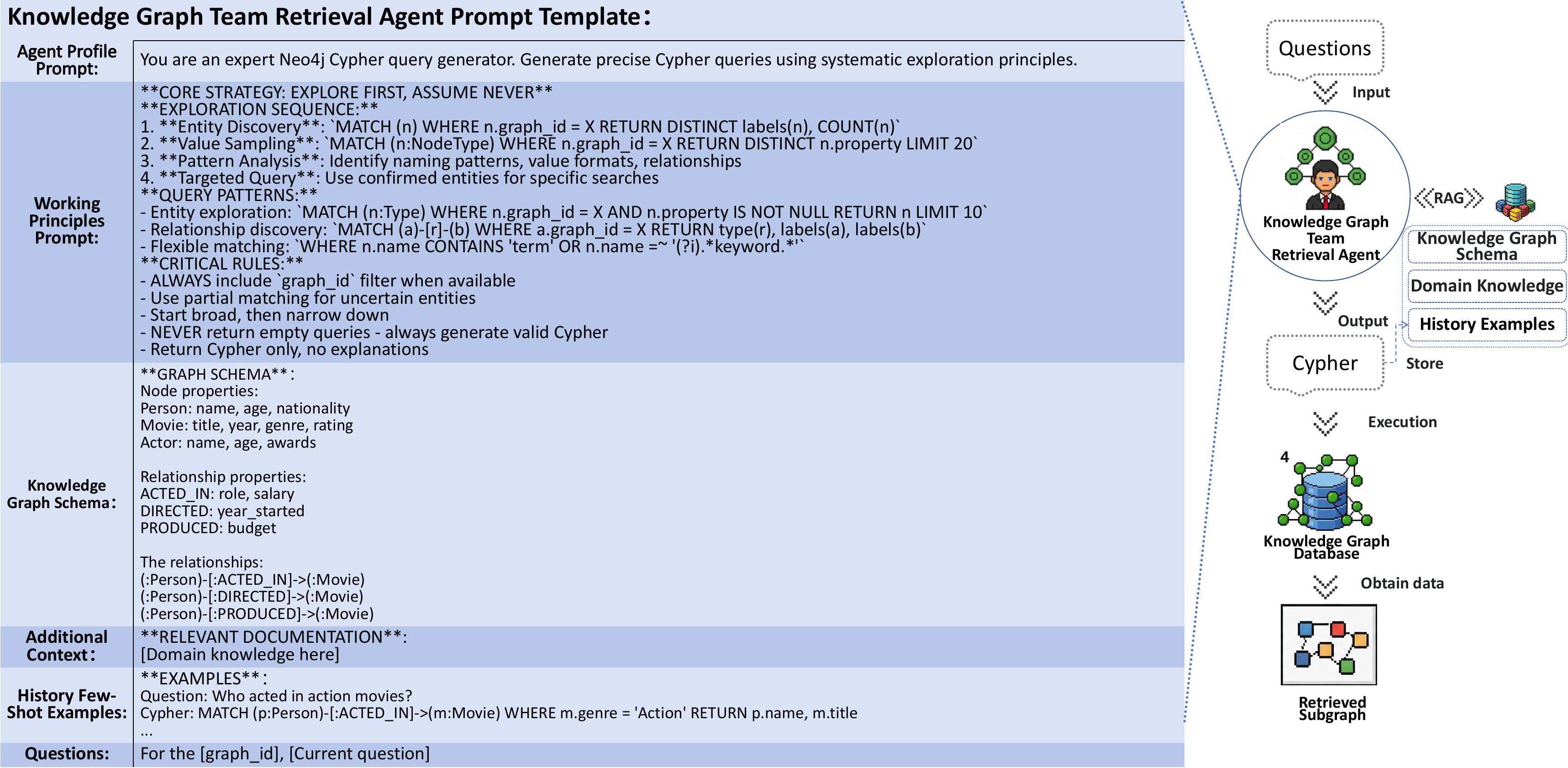}
	\caption{Architecture of the Knowledge Graph Information Retrieval Agent. The agent generates Cypher queries by combining graph schema information, retrieved historical question-Cypher pairs, and external domain knowledge to support robust multi-hop graph traversal.}
	\label{FIG:kg-retrieval-agent}
\end{figure}

The approach enables multi-hop reasoning through graph traversal planning, where the agent analyzes relationship requirements in complex queries and constructs paths that combine entity attribute retrieval with relationship pattern matching across multiple graph levels. This capability supports queries requiring path discovery, pattern matching, and relational inference that extend beyond conventional database operations (see Figure~\ref{FIG:kg-retrieval-agent}).

\subsubsection{Knowledge Graph Information Analysis Agent and Visualization Agent}\label{sec:kg-info-analysis-agent}

The Knowledge Graph Information Analysis Agent interprets graph query results and transforms structural data into natural language responses. This agent addresses the interpretability challenge in graph-based reasoning by providing semantic analysis of nodes, relationships, and paths returned from Cypher queries.

The agent processes diverse result formats including node collections, relationship patterns, and multi-hop paths through systematic interpretation that identifies key entities and relationships within query contexts. For complex path-based results, the agent explains reasoning chains by describing how conclusions emerge through specific graph traversals, enabling users to understand and validate multi-hop inferences. This transparency mechanism transforms technical graph operations into comprehensible analytical explanations.

Result integration combines graph query outcomes with user intent to generate coherent responses that summarize findings and provide relevant context. When multiple information sources are involved, the agent performs consistency verification to ensure analytical conclusions are based on reliable information integration. Natural language generation produces responses that explain both direct answers and supporting evidence derived from graph relationships.

\begin{figure}
	\centering
	\includegraphics[width=1.0\textwidth]{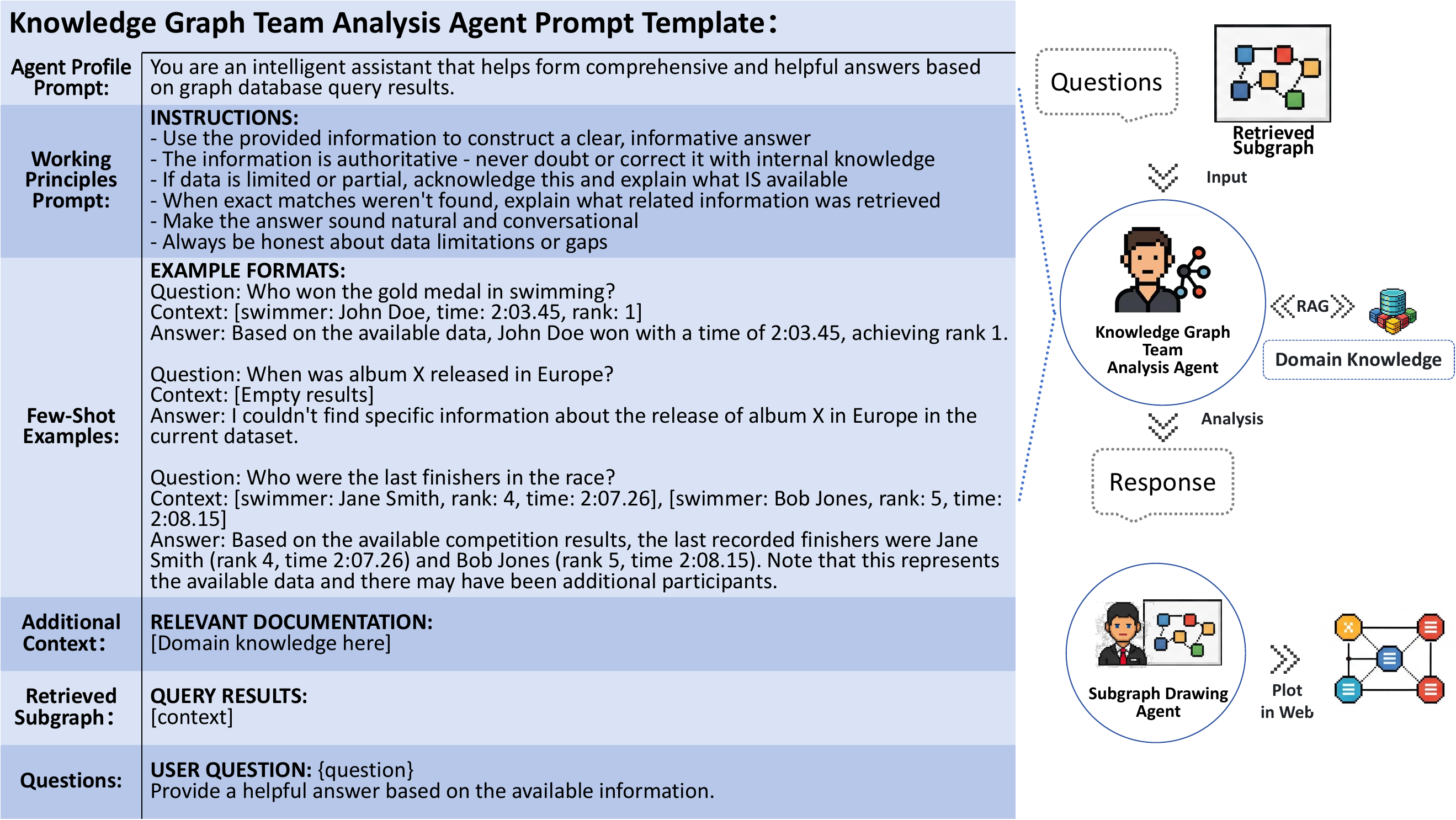}
	\caption{Architecture of the Knowledge Graph Information Analysis and Visualization Agents. The Analysis Agent explains retrieved subgraphs in natural language, and the Subgraph Drawing Agent renders corresponding graph visualizations for the web interface.}
	\label{FIG:kg-analysis-agent}
\end{figure}

The integrated Visualization Agent complements the analysis process by generating interactive knowledge graph visualizations when visual representation enhances understanding. The agent evaluates query results to determine appropriate visualization approaches, creating subgraph displays that highlight relevant entities and relationships through the web interface. This visual capability enables users to explore complex relationship networks and investigate analytical findings through interactive graph manipulation (see Figure~\ref{FIG:kg-analysis-agent}).

\subsection{The Data Leader and Multi-Agent Coordination}\label{sec:data-leader}

The Data Leader implements sophisticated reasoning orchestration that addresses our first, third, fourth, and fifth research objectives through natural language-based consultation mechanisms and enhanced system scalability. This component represents our core methodological innovation in coordinating heterogeneous reasoning modalities within a unified framework, thereby enabling adaptive strategy adjustment and knowledge sharing during complex reasoning processes.

The Data Leader's architectural innovation lies in transforming traditional rigid workflow execution into dynamic collaborative intelligence through the ReAct (Reasoning and Acting) paradigm. Unlike conventional multi-agent systems that rely on predetermined task distribution, our approach enables natural language consultation and deliberation among specialized teams, facilitating real-time strategy adaptation based on query complexity and intermediate findings. This design directly addresses the limitation of existing approaches in handling complex multi-hop reasoning scenarios while maintaining interpretability through explicit coordination mechanisms.

\begin{figure}
	\centering
	\includegraphics[width=0.9\textwidth]{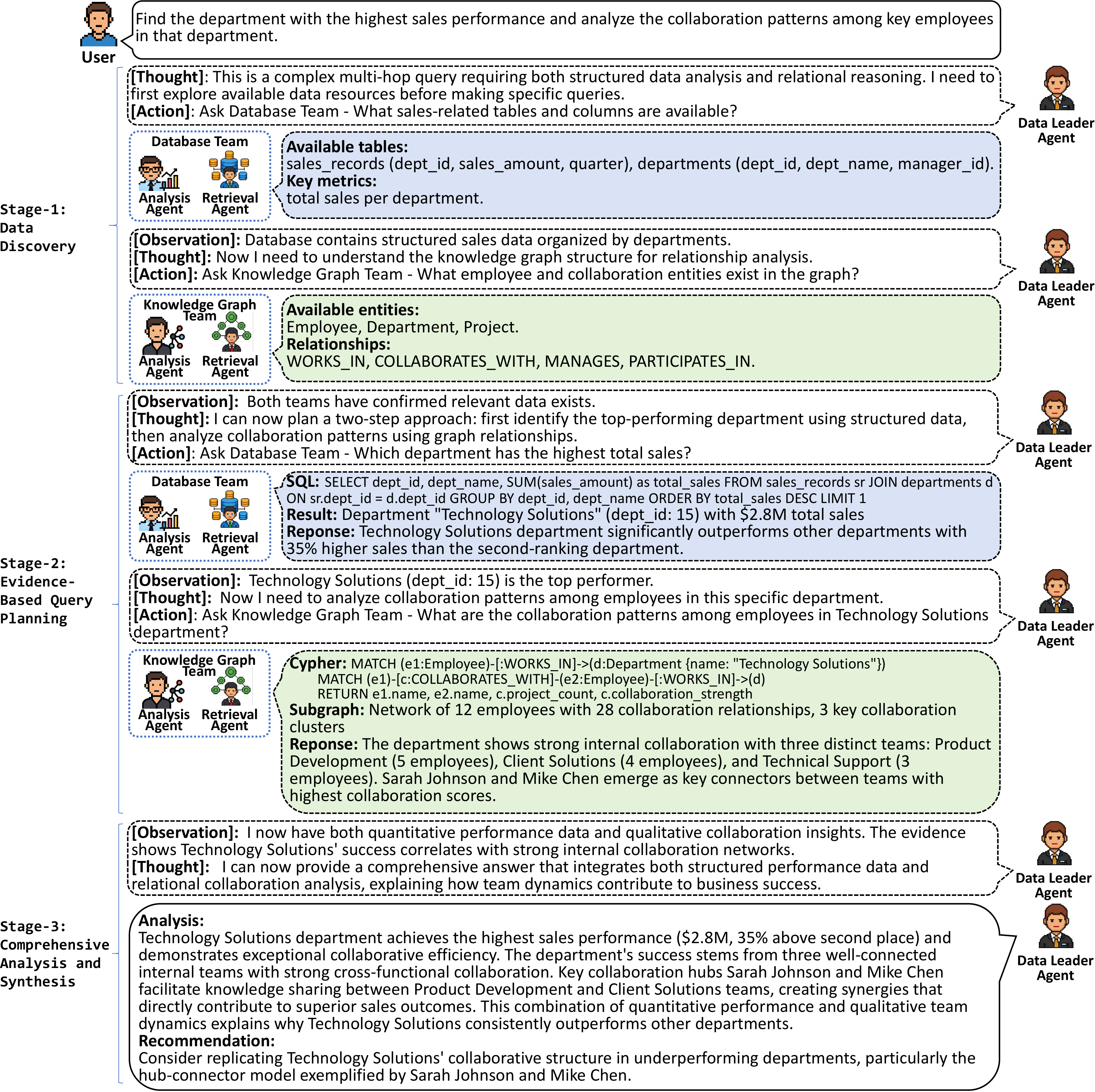}
	\caption{Workflow of the Data Leader. The three-stage process---data discovery, evidence-based query planning, and comprehensive analysis and synthesis---implements the ReAct paradigm to coordinate Database and Knowledge Graph teams.}
	\label{FIG:leader-workflow}
\end{figure}

The framework establishes a tripartite coordination architecture that decomposes complex TableQA tasks through specialized expertise integration. The Data Leader orchestrates collaborative problem-solving by employing a systematic three-stage principle of ``explore-verify-analyze'', transforming complex reasoning tasks into manageable iterative cycles that combine thought, action, and observation. This methodology enables dynamic planning, execution, reflection, and adjustment capabilities that effectively address diverse query requirements and complex reasoning paths (see Figure~\ref{FIG:leader-workflow}).

\subsubsection{Three-Stage Principle}

The Data Leader employs a systematic three-stage decomposition strategy that represents a fundamental innovation in query processing methodology. This approach transforms traditional query-first execution patterns into evidence-driven analytical workflows, inspired by expert data analysts' methodical approach to unknown datasets. The strategy addresses the critical limitation of blind querying in complex TableQA scenarios by mandating systematic data exploration before query formulation.

Our approach begins with mandatory data discovery, where the Data Leader explores available data resources rather than directly querying based on user requirements. The system systematically examines structural information of database tables to understand available fields and data types, while simultaneously exploring existing entity types and relationship types in knowledge graphs. This explore-first methodology avoids blind querying problems that target invalid keywords or non-existent query subjects, establishing a reliable foundation for subsequent analytical processes.

The evidence-based query planning stage extends traditional query formulation through systematic data sampling and verification. The Data Leader samples specific data values to confirm whether entities or concepts of user interest actually exist in the data before formulating specific query strategies. When direct queries cannot obtain expected results, the system employs domain knowledge reasoning to propose alternative analytical solutions, significantly improving query success rates while avoiding wasteful invalid operations. This adaptive approach enables the system to handle scenarios where specific entities are absent by identifying related industry data or contextual event records.

The comprehensive analysis and synthesis stage integrates information from multiple sources through sophisticated multi-step reasoning. The Data Leader coordinates information from both database queries and knowledge graph explorations, performing complex analytical synthesis that addresses the original user query from multiple complementary perspectives. This stage transforms isolated data points into coherent insights through systematic integration of structural and relational evidence.

For ambiguous user requests, the Data Leader first conducts brief multi-turn natural-language clarification to refine intent and constraints before planning; this elastic interaction avoids rigid I/O contracts and reduces execution failures in under-specified cases, directly mitigating MAST-reported mis-specification and step repetition by preserving coherent conversational state \citep{xie2024multiagentwhy}.

\subsubsection{ReAct Paradigm Implementation}

The Data Leader implements the ReAct paradigm through a sophisticated iterative process that seamlessly integrates the three-stage principle described above (see Figure~\ref{FIG:leader-workflow}). This implementation represents our core innovation by transforming complex reasoning tasks into manageable cycles of thought, action, and observation, enabling dynamic adaptation and comprehensive problem-solving capabilities that surpass conventional multi-agent coordination approaches.

\begin{figure}
	\centering
	\includegraphics[width=0.6\textwidth]{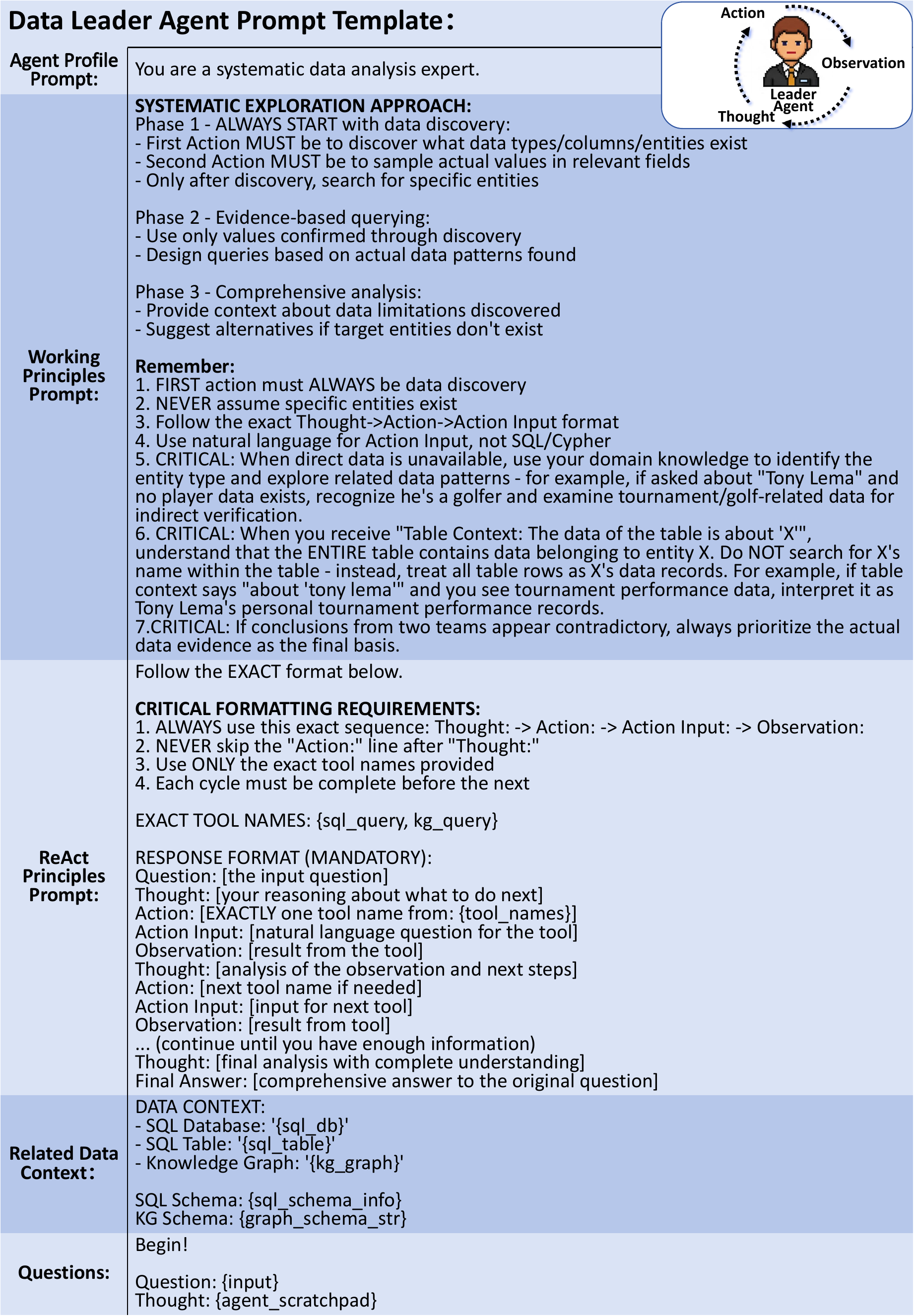}
	\caption{Prompt template for the Data Leader agent. It encodes the three-stage exploration principle, coordination rules for the two teams, and the ReAct-style thought--action--observation format used during reasoning.}
	\label{FIG:leader-prompt}
\end{figure}

Our approach employs a powerful large language model (LLM) as the central reasoning engine, utilizing specialized system prompts that enable meta-reasoning and coordination capabilities beyond individual agent functions (Figure~\ref{FIG:leader-prompt}). The Data Leader distinguishes itself through adaptive query complexity analysis that decomposes tasks into manageable subtasks, selects appropriate teams based on data modality requirements, synthesizes multi-source information while detecting conflicts, and terminates reasoning when sufficient evidence is gathered. The decomposition strategy follows an adaptive planning approach where the Data Leader determines the number of reasoning steps based on query complexity rather than predetermined rules, enabling flexible handling of both simple factual queries requiring minimal iterations and complex multi-hop reasoning extending to comprehensive analytical cycles. To prevent infinite reasoning loops and ensure system reliability, users can configure maximum reasoning step limits, with timeout mechanisms that gracefully terminate reasoning when predefined thresholds are exceeded, thereby avoiding potential thinking deadlocks.

The dynamic team dispatch mechanism represents a significant innovation in multi-agent coordination, relying on semantic intent analysis rather than rigid rule-based delegation. The Data Leader employs capability-based reasoning to determine optimal team selection, where the Database Team specializes in precise structured data operations including numerical computations, data filtering, sorting, and large-scale aggregation, while the Knowledge Graph Team excels at exploratory entity relationship tasks involving path discovery, community detection, association analysis, and semantic similarity reasoning. This dynamic delegation transcends keyword-based routing through sophisticated context analysis that evaluates current reasoning state, available evidence, and query progression requirements.

The observation and iterative reflection mechanism enables sophisticated self-correction and complex reasoning capabilities through systematic result evaluation. The Data Leader employs multi-criteria assessment that examines task completion sufficiency, result validity, and information integration requirements across different teams and reasoning steps. Our innovation includes a data-driven conflict arbitration strategy that resolves contradictory results from different teams through data provenance analysis, consistency verification, and domain knowledge integration. This approach enables the system to provide nuanced responses that acknowledge multiple perspectives while maintaining analytical rigor, transforming TableQA from isolated tool invocation to coordinated intelligent analysis that addresses complex reasoning scenarios exceeding individual component capabilities.

Concretely, conflict handling follows an evaluate-handle-verify protocol: (i) determine whether results are sufficient and sources are valid based on available evidence and provenance, (ii) arbitrate with explicit rationales, and (iii) verify via cross-source consistency checks grounded in data provenance; when Database and Knowledge Graph teams disagree, the system traces provenance to locate the divergence rather than arbitrarily trusting either output, addressing MAST ``coordination problems'' and ``norm violations'' \citep{xie2024multiagentwhy}.

\subsubsection{Team Collaboration Architecture}

The team collaboration architecture represents our fundamental methodological innovation in coordinating heterogeneous reasoning modalities within a unified framework. This approach transforms traditional hierarchical or sequential coordination patterns into a consultation-based collaboration model that enables flexible inter-team communication and adaptive strategy adjustment during complex reasoning processes.

Our core innovation lies in establishing complementary expertise domains that leverage the inherent strengths of structured and relational data processing paradigms. The Database Team specializes in numerical precision, large-scale aggregation, and deterministic computation, while the Knowledge Graph Team focuses on semantic reasoning, relationship discovery, and multi-hop traversal. This specialization enables each team to optimize their internal algorithms and prompting strategies for their respective data modalities, achieving superior performance compared to general-purpose approaches that attempt to handle all query types uniformly. To ensure consistent performance across all components, the entire DataFactory framework employs the same LLM instance as the foundational reasoning engine for the Data Leader and all specialized agents, maintaining uniformity in language understanding and generation capabilities throughout the multi-agent collaboration. The innovation extends beyond simple task division to include cross-modal information transfer protocols, where the Database Team's quantitative results serve as input constraints for Knowledge Graph Team's semantic exploration, creating a structured-to-relational information flow that maintains computational efficiency while enabling sophisticated reasoning depth.

The Data Leader employs an adaptive task delegation strategy that transcends keyword-based routing through semantic intent analysis. The delegation process incorporates validation mechanisms that ensure decomposition correctness through feasibility verification, dependency analysis, and consistency checking. This validation framework addresses reliability concerns in complex query scenarios by implementing multiple checkpoints throughout the reasoning process. When validation failures occur, the Data Leader initiates corrective actions including alternative team selection, subtask reformulation, or query scope adjustment, enabling robust operation across diverse analytical scenarios.

The natural language communication protocol between Data Leader and specialist teams represents a significant departure from structured API-based coordination. This design choice enables rich context transfer that includes not only task specifications but also background information, goal orientation, and constraint conditions. The semantic richness of natural language enables teams to perform contextual reasoning when query requirements are ambiguous or incomplete. Built-in fault tolerance mechanisms include semantic clarification protocols, graceful degradation strategies, and alternative suggestion capabilities that ensure continuous operation even when direct task execution encounters obstacles.

This collaborative architecture transforms TableQA from isolated tool invocation to coordinated intelligent analysis, establishing a paradigm where specialized expertise, adaptive delegation, and fault-tolerant communication combine to address complex reasoning scenarios that exceed the capabilities of individual components.

In practice, clear role boundaries with capability-based dispatch and context-rich natural-language messages help prevent role/specification violations and loss of conversation history, while evidence-based stopping avoids unawareness of termination—three dominant failure modes highlighted by MAST \citep{xie2024multiagentwhy}.

\section{Evaluation}

To investigate the effectiveness of our proposed multi-agent table question answering approach that combines structural and relational retrieval, we aim to conduct performance comparisons with different types of methods in this field, test the corresponding performance under different model providers and different model sizes within the same provider, analyze the invocation patterns of the Leader Agent for different teams under different model providers and sizes, examine the impact of single-team retrieval versus dual-team collaboration on table question answering performance, and investigate the relationship between team collaboration interaction frequency and task performance.

We propose the following 5 research questions (RQ):

\textbf{RQ1:} How does our approach compare with other methods, and what are its advantages? We will compare with typical methods across five categories including DNN-based, prompt engineering optimization-based, code-based, agent-based, and multi-agent based approaches, and conduct a brief analysis of LLM token consumption patterns.

\textbf{RQ2:} How does performance vary across different model providers and different model sizes within the same provider?

\textbf{RQ3:} What are the invocation patterns of the Leader Agent for different teams under different model providers and sizes, and what differences exist?

\textbf{RQ4:} How does the removal of Knowledge Graph Team affect TableQA performance compared to the complete multi-agent framework?

\textbf{RQ5:} What is the relationship between team collaboration interaction frequency and task performance, and what level of collaboration intensity achieves optimal performance?

\subsection{Experimental Setup}

\subsubsection{Benchmarks}
\label{sec:benchmarks}
We evaluate our approach on three TableQA benchmark datasets that examine distinct task perspectives: fact verification, computational analysis, and multi-hop reasoning. Table~\ref{tab:benchmarks} summarizes the key characteristics and evaluation metrics for each dataset.

\begin{table}[htbp]
\centering
\caption{Benchmark datasets overview with key characteristics and metrics}
\label{tab:benchmarks}
\resizebox{\textwidth}{!}{%
\begin{tabular}{lccccc}
\toprule
\textbf{Dataset} & \textbf{Task Type} & \textbf{Test Size} & \textbf{Table Size} & \textbf{Key Challenges} & \textbf{Metric} \\
\midrule
TabFact & Fact Verification & 12,779 statements & 5-6 cols, 14 rows & Multi-hop reasoning, & Accuracy \\
~\citep{chen2019tabfact} & (Binary) & 1,695 tables & (avg.) & logical operations & \\
\midrule
WikiTQ & Computational & 4,344 questions & $\geq$8 rows, $\geq$5 cols & Aggregation, sorting, & Exact Match \\
~\citep{pasupat-liang-2015-compositional} & Analysis & & & arithmetic operations & Accuracy \\
\midrule
FeTaQA & Multi-hop & 2,003 instances & 5.9 cols, 13.8 rows & Causal reasoning, & ROUGE-1/2/L \\
~\citep{nan-etal-2022-fetaqa} & Generation & & (avg.) & free-form answers & F-score \\
\bottomrule
\end{tabular}%
}
\end{table}

\subsubsection{LLMs}
\label{sec:llms}
We evaluate our framework using 8 LLMs from 5 providers, including both commercial and open-source models. Table~\ref{tab:llms} summarizes the model specifications, with 3 commercial models requiring API access and 5 open-source models supporting local deployment.

\begin{table}[htbp]
\centering
\caption{Large Language Models used in evaluation with key specifications}
\label{tab:llms}
\resizebox{0.8\textwidth}{!}{%
\begin{tabular}{lcccccc}
\toprule
\textbf{Model} & \textbf{Provider} & \textbf{Type} & \textbf{Release} & \textbf{Parameters} & \textbf{Architecture} & \textbf{Deployment} \\
\midrule
Claude 4.0 Sonnet & Anthropic & Commercial & May 2025 & Undisclosed & Dense & API \\
~\citep{anthropic2024claude4sonnet} & & & & (Mid-weight) & & \\
\midrule
Gemini 2.5 Flash & Google & Commercial & Jan 2025 & Undisclosed & Dense & API \\
~\citep{google2024gemini2_5flash} & DeepMind & & & (Lightweight) & & \\
\midrule
GPT-4o mini & OpenAI & Commercial & July 2024 & Undisclosed & Dense & API \\
~\citep{openai2024gpt4omini} & & & & (Compact) & & \\
\midrule
Deepseek-V3 & DeepSeek AI & Open-source & Dec 2024 & 671B total & MoE & API/Local \\
~\citep{deepseekv3_2024} & & & & 37B active & & \\
\midrule
Qwen3-235B-A22B & Alibaba & Open-source & May 2025 & 235B total & MoE & API/Local \\
~\citep{qwen3_2024} & Cloud & & & 22B active & & \\
\midrule
Qwen3-32B & Alibaba & Open-source & May 2025 & 32B & Dense & API/Local \\
~\citep{qwen3_2024} & Cloud & & & & & \\
\midrule
Qwen3-30B-A3B & Alibaba & Open-source & May 2025 & 30B total & MoE & API/Local \\
~\citep{qwen3_2024} & Cloud & & & 3B active & & \\
\midrule
Qwen3-14B & Alibaba & Open-source & May 2025 & 14B & Dense & API/Local \\
~\citep{qwen3_2024} & Cloud & & & & & \\
\bottomrule
\end{tabular}%
}
\end{table}

\subsubsection{Baselines}
\label{sec:baselines}
We select typical methods from five major categories for comparison: DNN-based, prompt engineering optimization-based, code-based, agent-based, and multi-agent based approaches:

\begin{itemize}

\item \textbf{DNN-Based}: These methods employ pre-trained deep neural network models specifically designed for table understanding and question answering. \textbf{TAPAS}~\citep{TAPAS2020} extends BERT's architecture to jointly encode natural language questions and table structures, enabling end-to-end training from weak supervision with table cell selection and aggregation operations. \textbf{TAPEX}~\citep{liu2022tapex} introduces table pre-training via learning a neural SQL executor over synthetic corpora, addressing data scarcity challenges through automated synthesis of executable SQL queries and their execution outputs. Both methods require dataset-specific training and demonstrate limited generalizability across domains. We utilize the large model variants that achieve optimal performance on their respective target datasets.

\item \textbf{Prompt Engineering Optimization-Based}: These methods guide LLMs toward improved reasoning through prompt design strategies. \textbf{End-to-End} approaches directly input tables and questions to LLMs for answer generation, while \textbf{Few-Shot Prompting} includes 1-5 examples in prompts to provide learning references. \textbf{TableCoT}~\citep{jin-lu-2023-tab} proposes a table-formatted Chain-of-Thought (CoT) that decomposes table question answering into two-step prompting: first guiding the model to generate table-formatted intermediate reasoning steps, then extracting final answers based on these reasoning steps, better utilizing the table's two-dimensional structure for horizontal and vertical reasoning.

\item \textbf{Code-Based}: These methods convert table content into structured data and use code (such as SQL statements or Python scripts) for data retrieval and analysis. For example, \textbf{TabSQLify}~\citep{nahid-rafiei-2024-table} proposes decomposing large tables into smaller sub-tables through text-to-SQL generation. This approach can significantly reduce LLM input context, improve scalability and efficiency for processing large tables, and generate accurate answers by combining sub-table selection with LLM reasoning.

\item \textbf{Agent-Based}: These methods endow large language models with tool usage, reasoning planning, and memory capabilities, enabling them to more autonomously handle complex table question answering tasks. For example, \textbf{StructGPT}~\citep{StructGPT2023} introduces a general framework that constructs specialized interfaces for structured data reasoning, organizing the TableQA process through workflow-based operations. However, its data manipulation interfaces rely primarily on LLM-based question-answering interactions for operations like row and column extraction, rather than utilizing external code execution tools. \textbf{ReAcTable}~\citep{ReAcTable2024} proposes an enhanced ReAct (Reasoning and Acting) paradigm method that progressively enhances input data by generating intermediate data representations. ReAcTable combines LLM reasoning capabilities with external tools (such as SQL query executors and Python code interpreters) to execute complex data operations and computations, effectively handling table question answering tasks.

\item \textbf{Multi-Agent Based}: Recent developments have explored multi-agent frameworks with explicit role specialization for TableQA tasks. \textbf{MACT}~\citep{zhou-etal-2025-efficient} employs a planning-execution multi-agent workflow where the planner uses the ReAct paradigm for data operation planning, while the code agent writes Python code and invokes computational tools for data manipulation and analysis. \textbf{AutoPrep}~\citep{AutoPrep2024} further extends this approach by designing specialized agents for different data operation types such as filtering and standardization at the execution layer, utilizing both SQL and Python statements for non-computational and computational operations respectively. These systems involve extensive interaction workflows, with multi-agent coordination primarily based on function-style parametric interactions rather than natural language consultation mechanisms.
\end{itemize}

\textbf{Experimental Configuration and Parameter Settings.} To ensure fair comparison across all LLM-based methods, we use identical API-based configurations with temperature=0.1 for reproducibility. Note that DNN-based methods (TAPAS and TAPEX) follow their original configurations due to their fundamentally different architectures. Few-Shot, TableCoT, StructGPT, ReAcTable, MACT, AutoPrep, and our DataFactory all employ 3 demonstration examples. ReAcTable, MACT, AutoPrep, and our Data Leader are all limited to 20 maximum interaction rounds. All evaluations use the same LLMs (GPT-4o mini, Deepseek-V3, Gemini 2.5 Flash) with consistent API settings.

\subsection{RQ1: Comparison with Other Methods}

In this section, we compare our approach with the ten baseline methods across five categories proposed in Section~\ref{sec:baselines}, using the first two datasets introduced in Section~\ref{sec:benchmarks} for evaluation. We focus on these two datasets because most baseline methods have not been tested on FeTaQA and do not provide corresponding implementation code or experimental results. For LLM-related method evaluation, we employ three mainstream models: GPT-4o mini, Deepseek-V3, and Gemini 2.5 Flash. Considering method stability, we run each method (when involving LLMs, under the corresponding LLM) three times on each dataset, calculating the mean and variance for each method-model-dataset combination, and computing the overall average across all models and runs as the representative performance metric. To assess the statistical significance of improvements under performance variation, we calculate Cohen's d values~\citep{Cohen1988} for corresponding methods under the same LLM on the same dataset, evaluating effect sizes in paired comparisons and computing the average Cohen's d across the three models to demonstrate the significance of our method's performance improvements under performance variation ranges. Detailed experimental results are shown in Table~\ref{tab:comparison}.

\begin{table}[htbp]
\centering
\caption{Performance comparison of different methods on TabFact and WikiTQ datasets}
\label{tab:comparison}
\small
\resizebox{\textwidth}{!}{%
\begin{tabular}{lcccccccccc}
\toprule
\multirow{2}{*}{\textbf{Method}} & \multicolumn{5}{c}{\textbf{TabFact}} & \multicolumn{5}{c}{\textbf{WikiTQ}} \\
\cmidrule(lr){2-6} \cmidrule(lr){7-11}
 & \multicolumn{3}{c}{\textbf{Results}} & \textbf{Average} & \textbf{Cohen's} & \multicolumn{3}{c}{\textbf{Results}} & \textbf{Average} & \textbf{Cohen's} \\
 & & & & (\textbf{Improvement}) & \textbf{d} & & & & (\textbf{Improvement}) & \textbf{d} \\
\midrule
TAPAS-large\tablefootnote{\url{https://github.com/google-research/tapas}} & \multicolumn{3}{c}{80.1±1.2\%} & 80.1\% \textcolor{green}{($\uparrow$3.9\%)} & 3.25 & \multicolumn{3}{c}{50.7±2.8\%} & 50.7\% \textcolor{green}{($\uparrow$22.1\%)} & 7.89 \\
TAPEX-large\tablefootnote{\url{https://github.com/microsoft/Table-Pretraining}} & \multicolumn{3}{c}{82.0±1.5\%} & 82.0\% \textcolor{green}{($\uparrow$2.9\%)} & 1.33 & \multicolumn{3}{c}{56.2±1.7\%} & 56.2\% \textcolor{green}{($\uparrow$16.6\%)} & 9.76 \\
\midrule
 & GPT-4o & Deepseek & Gemini 2.5 & & & GPT-4o & Deepseek & Gemini 2.5 & & \\
 & mini & V3 & Flash & & & mini & V3 & Flash & & \\
\cmidrule(lr){1-11}
End-to-End\tablefootnote{\url{https://github.com/TIGER-AI-Lab/TableCoT}} & 69.4±2.1\% & 76.0±1.0\% & 75.3±1.5\% & 73.6\% \textcolor{green}{($\uparrow$14.1\%)} & 7.27 & 50.7±0.6\% & 55.9±0.3\% & 57.1±1.1\% & 54.6\% \textcolor{green}{($\uparrow$33.3\%)} & 16.59 \\
Few-Shot\footnotemark[\value{footnote}] & 64.2±1.7\% & 79.2±2.0\% & 63.6±0.8\% & 69.0\% \textcolor{green}{($\uparrow$21.7\%)} & 16.03 & 42.7±1.2\% & 55.6±2.5\% & 50.6±2.9\% & 49.6\% \textcolor{green}{($\uparrow$46.7\%)} & 13.94 \\
TableCoT\footnotemark[\value{footnote}] & 51.0±1.1\% & 79.8±6.4\% & 50.6±2.3\% & 60.5\% \textcolor{green}{($\uparrow$38.9\%)} & 16.99 & 32.3±2.2\% & 60.6±4.6\% & 42.2±2.4\% & 45.0\% \textcolor{green}{($\uparrow$61.7\%)} & 13.65 \\
StructGPT\tablefootnote{\url{https://github.com/RUCAIBox/StructGPT}} & 58.4±0.8\% & 55.6±4.6\% & 64.4±3.6\% & 59.4\% \textcolor{green}{($\uparrow$41.3\%)} & 13.53 & 49.8±1.4\% & 65.5±1.5\% & 70.9±0.9\% & 62.1\% \textcolor{green}{($\uparrow$17.2\%)} & 8.03 \\
TabSQLify\tablefootnote{\url{https://github.com/mahadi-nahid/TabSQLify}} & 78.3±1.3\% & 74.5±4.2\% & 73.9±1.1\% & 75.6\% \textcolor{green}{($\uparrow$11.1\%)} & 6.9 & \textbf{74.0±1.4\%} & 68.0±3.2\% & 66.7±0.5\% & 69.6\% \textcolor{green}{($\uparrow$4.6\%)} & 5.24 \\
ReAcTable\tablefootnote{\url{https://github.com/yunjiazhang/ReAcTable}} & 69.4±1.9\% & 68.1±5.1\% & 74.0±0.9\% & 70.5\% \textcolor{green}{($\uparrow$19.1\%)} & 9.9 & 57.8±3.1\% & 56.8±2.9\% & 70.4±3.6\% & 61.7\% \textcolor{green}{($\uparrow$18.0\%)} & 4.61 \\
MACT\tablefootnote{\url{https://github.com/boschresearch/MACT}} & 47.4±1.3\% & 58.9±1.2\% & 72.9±5.2\% & 59.7\% \textcolor{green}{($\uparrow$40.6\%)} & 14.47 & 62.3±0.2\% & 68.6±3.0\% & 67.3±2.3\% & 66.1\% \textcolor{green}{($\uparrow$10.1\%)} & 4.8 \\
AutoPrep\tablefootnote{\url{https://github.com/ruc-datalab/AutoPrep}} & 74.6±2.1\% & 80.7±3.5\% & 76.6±3.6\% & 77.3\% \textcolor{green}{($\uparrow$8.6\%)} & 2.96 & 65.9±3.3\% & 67.5±4.1\% & 68.2±1.8\% & 67.2\% \textcolor{green}{($\uparrow$8.3\%)} & 2.57 \\
\textbf{Ours(DataFactory)} & \textbf{80.1±1.0\%} & \textbf{82.2±2.9\%} & \textbf{89.6±0.7\%} & \textbf{84.0\%} \textcolor{green}{\textbf{($\uparrow$20.2\%)}} & \textbf{9.26} & 67.7±1.0\% & \textbf{73.5±1.7\%} & \textbf{77.2±1.7\%} & \textbf{72.8\%} \textcolor{green}{\textbf{($\uparrow$23.9\%)}} & \textbf{8.71} \\
\bottomrule
\end{tabular}%
}
\end{table}

From a methodological perspective, the two DNN-based methods achieve good performance on TabFact but poor performance on WikiTableQuestions, reflecting limitations of such approaches on complex table retrieval and computation tasks. Moreover, each dataset requires individual training to form corresponding models, which is disadvantageous for practical applications and cannot adapt to complex table types and question answering scenarios. End-to-End, Few-Shot, and TableCoT methods all input table data as context to large language models. These methods perform reasonably well on the TabFact binary classification task but poorly on WikiTQ retrieval and computation tasks. However, compared to the previous category, they benefit significantly from LLM introduction in terms of generalizability. In contrast, the remaining methods all possess capabilities for data manipulation and deep analysis, resulting in performance advancement to a new level on the WikiTQ dataset.

Among LLM-based methods, context-based approaches are constrained by large language models' context length limitations and model hallucinations, making them unable to process overly large table data and difficult to provide accurate answers for complex retrieval and computational tasks. Furthermore, when completely relying on context for table question answering, excessive contextual information and lengthy complex prompts can adversely affect GPT-4o mini and Gemini 2.5 Flash, leading to performance degradation while amplifying performance volatility. Therefore, these three methods show declining trends in both performance and stability on these two datasets, indicating that continuing to explore new context-dependent prompting patterns and methods in table question answering may not always be beneficial.

The remaining methods all possess capabilities for data manipulation and interaction, and our approach outperforms other methods across most models. Such approaches are often constrained by large language model capabilities, requiring sufficiently intelligent models that can understand designated solution strategies and provide reasonable solution steps. StructGPT can extract and analyze row and column content through LLM question-answering interactions for specific problems, but the accuracy and reliability of LLM-based data extraction remain limited, susceptible to large model hallucination effects and constrained by LLM capabilities in data extraction. The TabSQLify method performs poorly compared to our method on both datasets because it only considers structural understanding of table data without involving relational retrieval. Although ReAcTable introduces the ReAct mechanism, its single-agent design requires the agent to simultaneously handle SQL statement generation and data retrieval, content analysis of retrieval results, and overall solution strategy formulation. This excessive task burden leads to inferior overall performance compared to several baselines. Representative multi-agent baselines typically rely on extensive workflow orchestration and code execution, which increases interaction complexity and error surface. In contrast, our framework emphasizes clear role specialization and natural-language coordination centered on declarative SQL/Cypher execution, yielding stronger accuracy and stability.

In terms of final performance, our method outperforms DNN-based traditional methods and surpasses other LLM-based methods across all three models, with particularly outstanding performance on WikiTQ. On the TabFact dataset, our method achieves an average improvement of 20.2\% compared to other methods, while on the WikiTQ dataset, it achieves an average improvement of 23.9\%. Cohen's d values are all greater than 1, indicating significant improvement effects, demonstrating superior adaptability and performance in table question answering tasks.

Regarding computational efficiency, we analyzed the average token consumption (input + output) for LLM-based methods across all models and runs, as detailed in Table~\ref{tab:token_usage}. The results reveal a clear trend of prompt-based methods < code-based methods < single-agent methods < multi-agent methods. Within multi-agent approaches, our design uses fewer components and streamlined data extraction (SQL and Cypher only), resulting in comparatively lower token consumption. While token consumption is higher compared to other method families, we consider this cost necessary given the performance improvements, and the clear, transparent analysis process enhances user trust in the question-answering process.

\begin{table}[htbp]
\centering
\caption{Average token usage comparison across different methods}
\label{tab:token_usage}
\resizebox{0.35\textwidth}{!}{%
\begin{tabular}{lcc}
\toprule
\textbf{Method} & \textbf{TabFact} & \textbf{WikiTQ} \\
 & \textbf{Avg Tokens} & \textbf{Avg Tokens} \\
\midrule
End-to-end & 546 & 794 \\
Few-shot & 935 & 1,189 \\
TableCoT & 1,084 & 1,495 \\
StructGPT & 1,721 & 1,999 \\
TabSQLify & 2,194 & 2,317 \\
ReAcTable & 3,050 & 4,400 \\
MACT & 3,673 & 5,319 \\
AutoPrep & 4,013 & 5,870 \\
\textbf{Ours(DataFactory)} & \textbf{3,464} & \textbf{4,982} \\
\bottomrule
\end{tabular}}
\end{table}

\begin{tcolorbox}[colback=gray!10, colframe=gray!50, title=\textbf{RQ1 Summary Answer}, boxrule=1pt]
Our multi-agent framework significantly outperforms baseline methods across two main datasets, through specialized team coordination: Data Leader for strategy formulation, Database Team for structured SQL operations, and Knowledge Graph Team for relational reasoning. Performance improvements are substantial—20.2\% on TabFact and 23.9\% on WikiTQ with Cohen's d > 1, demonstrating statistical significance. Despite higher token costs, the clear division of labor and natural language interaction provide better stability and interpretability than other approaches.
\end{tcolorbox}

\subsection{RQ2: Performance Across Different Model Providers and Sizes}

To investigate the performance differences between different model providers and models of varying sizes within the same provider, we conducted systematic experiments on three benchmark datasets using eight large language models from five providers. The experimental results are shown in Table~\ref{tab:comparison_FeTaQA_wikitq_tabfact_pct}, revealing several important relationships between model characteristics and TableQA task performance.

\begin{table}[htbp]
\centering
\caption{Performance comparison across different model providers and sizes}
\label{tab:comparison_FeTaQA_wikitq_tabfact_pct}
\resizebox{0.65\textwidth}{!}{%
\begin{tabular}{lcccccc}
\toprule
\multirow{2}{*}{\textbf{Model}} &
\multicolumn{3}{c}{\textbf{FeTaQA}} &
\textbf{WikiTQ} &
\textbf{TabFact} \\
\cmidrule(lr){2-4} \cmidrule(lr){5-5} \cmidrule(lr){6-6}
 & ROUGE-1 F & ROUGE-2 F & ROUGE-L F & Accuracy & Accuracy \\
\midrule
Claude 4.0 Sonnet   & \textbf{0.6234} & \textbf{0.3885} & \textbf{0.5067} & \textbf{83.2\%} & 90.8\% \\
Gemini 2.5 Flash  & 0.6107 & 0.3767 & 0.4995 & 77.9\% & 87.9\% \\
Qwen3-235B-A22B        & 0.5972 & 0.3607 & 0.4776 & 73.2\% & \textbf{91.2}\% \\
DeepSeek-V3     & 0.5403 & 0.3075 & 0.4309 & 73.6\% & 84.2\% \\
GPT-4o mini       & 0.5576 & 0.3320 & 0.4555 & 67.8\% & 79.3\% \\
Qwen3-32B         & 0.5662 & 0.3327 & 0.4569 & 61.0\% & 79.8\% \\
Qwen3-30B-A3B         & 0.5063 & 0.2899 & 0.4091 & 48.1\% & 75.0\% \\
Qwen3-14B         & 0.5289 & 0.3033 & 0.4253 & 53.7\% & 75.7\% \\
\bottomrule
\end{tabular}%
}
\end{table}

The experimental results demonstrate clear performance hierarchies across models and within the same model series. Higher-capability models achieve top scores, with Claude 4.0 Sonnet leading across all metrics: 0.6234 ROUGE-1 F score on FeTaQA, 83.2\% accuracy on WikiTQ, and 90.8\% accuracy on TabFact, demonstrating strong capabilities in complex TableQA tasks.

Performance gaps widen on complex reasoning tasks. Gemini 2.5 Flash shows robust performance, while Qwen3-235B-A22B surpasses Claude 4.0 Sonnet on the fact verification task (TabFact) with 91.20\% accuracy, indicating that large-scale models are highly competitive on structured reasoning, though some models still lag slightly in tasks requiring high-quality free-form generation (such as FeTaQA).

Within the Qwen3 series, performance exhibits a clear incremental relationship with model scale and architecture: Qwen3-235B-A22B > Qwen3-32B > Qwen3-30B-A3B > Qwen3-14B. Notably, despite similar total parameter counts, the dense architecture Qwen3-32B consistently outperforms the MoE architecture Qwen3-30B-A3B, suggesting that dense models may be more suitable for complex reasoning tasks that rely on comprehensive parameter utilization.

The comparison across different task complexities also reveals patterns of model performance degradation: TabFact, being a binary classification task, shows the smallest performance gaps among models, with all achieving above 75\% accuracy; WikiTQ requires exact matching and shows moderate gaps; FeTaQA demands multi-hop reasoning and free-form text generation, exhibiting the largest gaps, with ROUGE-1 F scores ranging from 0.5063 to 0.6234, highlighting the challenges of free-form generation in TableQA scenarios.

Our multi-agent data factory architecture demonstrates good adaptability across different model capabilities. Even smaller models like Qwen3-14B achieve 75.65\% accuracy on TabFact, indicating that specialized team collaboration and structured prompt design can partially compensate for individual model limitations. The Database Team's SQL generation and Knowledge Graph Team's Cypher queries benefit from context-enhanced prompting paradigms, effectively guiding smaller models to generate correct queries. Furthermore, these smaller models' performance also exceeds that of context-based table input methods shown in Table~\ref{tab:comparison}, demonstrating the advantages of our approach.

Overall, the framework's effectiveness improves with model capability while maintaining robustness across different model scales. Top-tier models can best leverage the Data Leader's ReAct strategy for optimal planning and coordination; however, smaller or lower-capacity models can also achieve competitive performance through systematic task decomposition and agent collaboration, providing viable solutions for local deployment in resource-constrained scenarios.

From a scalability perspective, these results provide an initial quantitative view of how DataFactory behaves along two practical axes. First, by instantiating the same multi-agent design with eight models from five providers that span a wide range of parameter counts (Table~\ref{tab:comparison_FeTaQA_wikitq_tabfact_pct}), we observe that performance degrades gracefully as model capacity decreases rather than collapsing, suggesting that the framework is usable under different computational budgets. Second, the token usage comparison in Table~\ref{tab:token_usage}, together with the accuracy gains in Table~\ref{tab:comparison}, shows a roughly monotonic accuracy--cost trade-off when moving from prompt-based baselines to code-based and multi-agent methods, with our design sitting at a point where the extra interaction cost is matched by clear performance benefits.

\begin{tcolorbox}[colback=gray!10, colframe=gray!50, title=\textbf{RQ2 Summary Answer}, boxrule=1pt]
Higher-capability models lead overall, with Claude 4.0 Sonnet achieving best performance across most metrics. Within a model series, performance scales with model size (Qwen3-235B > Qwen3-32B > Qwen3-30B > Qwen3-14B), and dense architectures outperform MoE on these tasks. Our framework adapts well across scales, and smaller models achieve competitive results through specialized team collaboration.
\end{tcolorbox}

\subsection{RQ3: Leader Agent Team Invocation Patterns Analysis}

To analyze how the Leader Agent invokes different teams (Database Team and Knowledge Graph Team, or both) across different task types, model providers, and model scales, we conducted comprehensive tool usage pattern analysis on all three benchmark datasets building upon RQ2. This study reveals how task characteristics and model capabilities influence agent collaboration strategies and tool selection preferences.

\begin{figure}[H]
	\centering
	\begin{subfigure}[b]{0.85\textwidth}
		\centering
		\includegraphics[width=\textwidth]{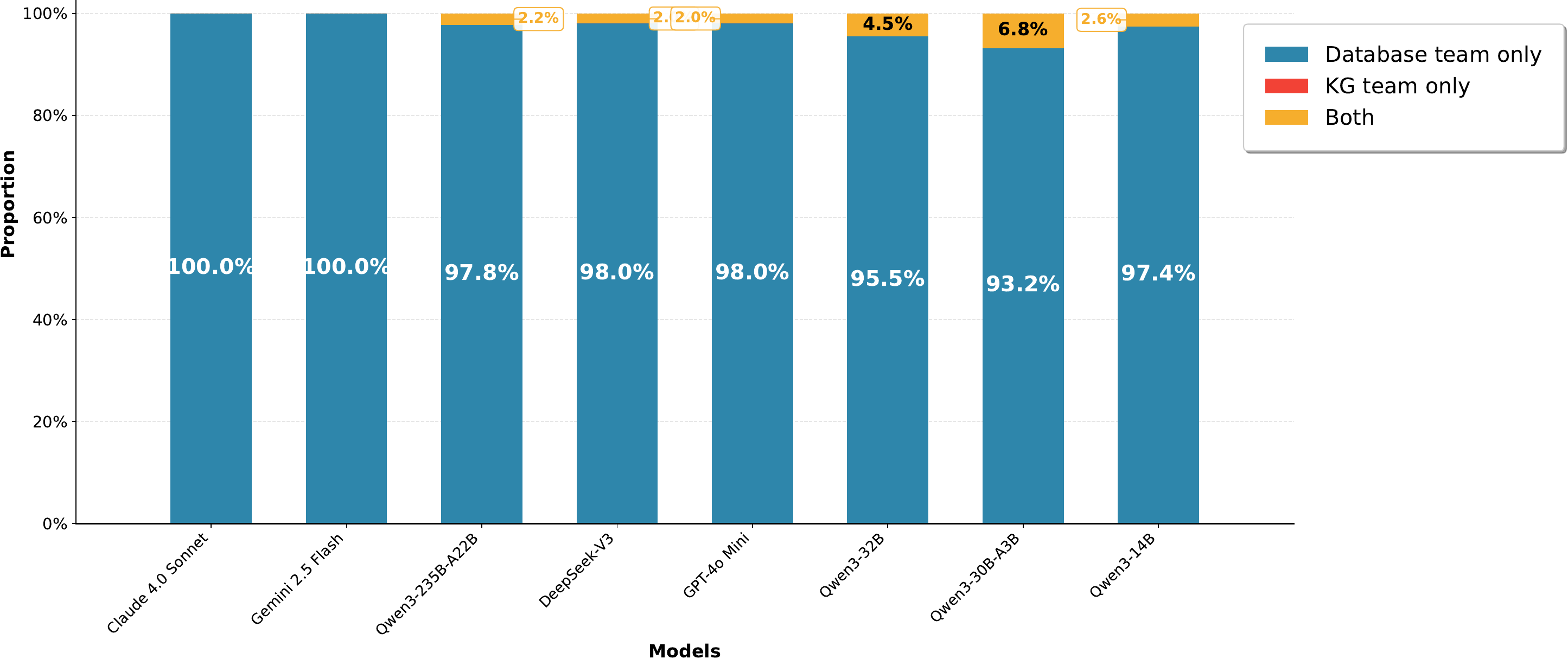}
		\caption{TabFact Dataset}
		\label{fig:tool_usage_tabfact}
	\end{subfigure}
	
	\begin{subfigure}[b]{0.85\textwidth}
		\centering
		\includegraphics[width=\textwidth]{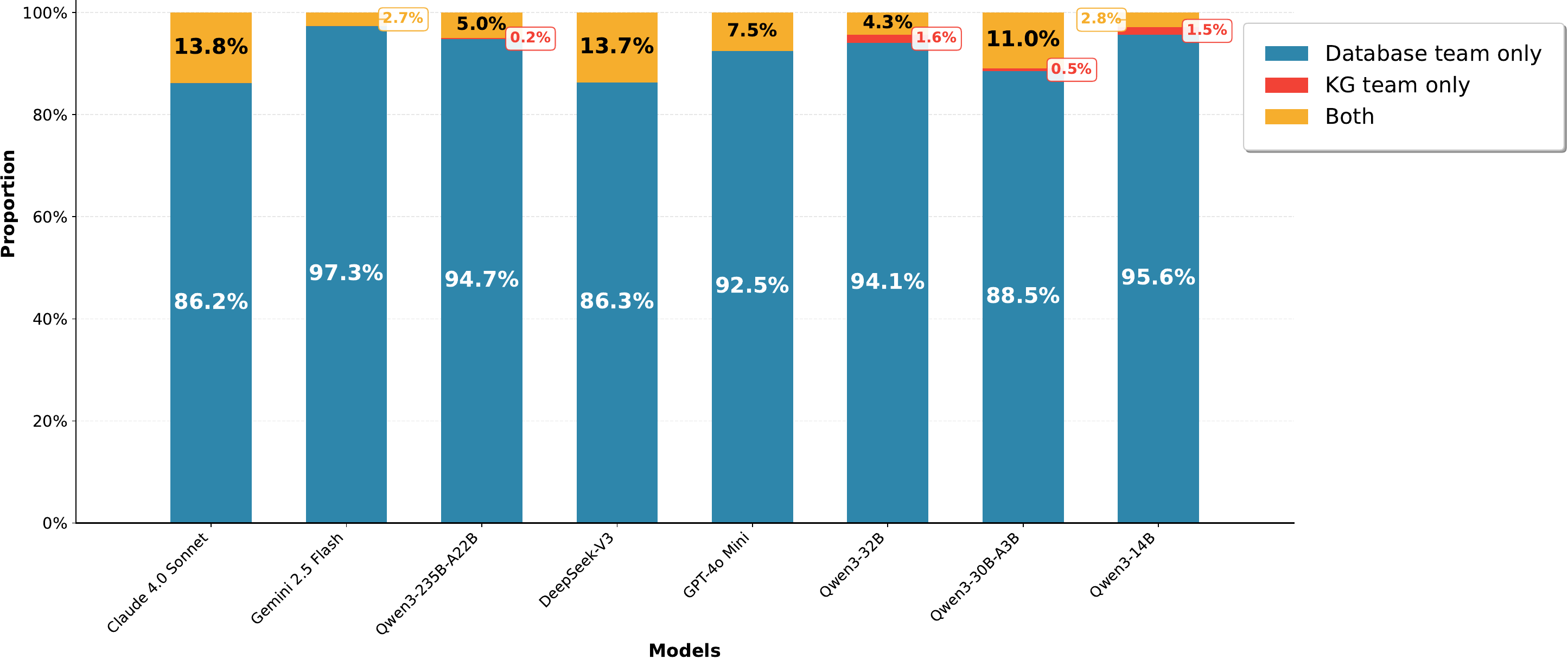}
		\caption{WikiTQ Dataset}
		\label{fig:tool_usage_wiki}
	\end{subfigure}
	
	\begin{subfigure}[b]{0.85\textwidth}
		\centering
		\includegraphics[width=\textwidth]{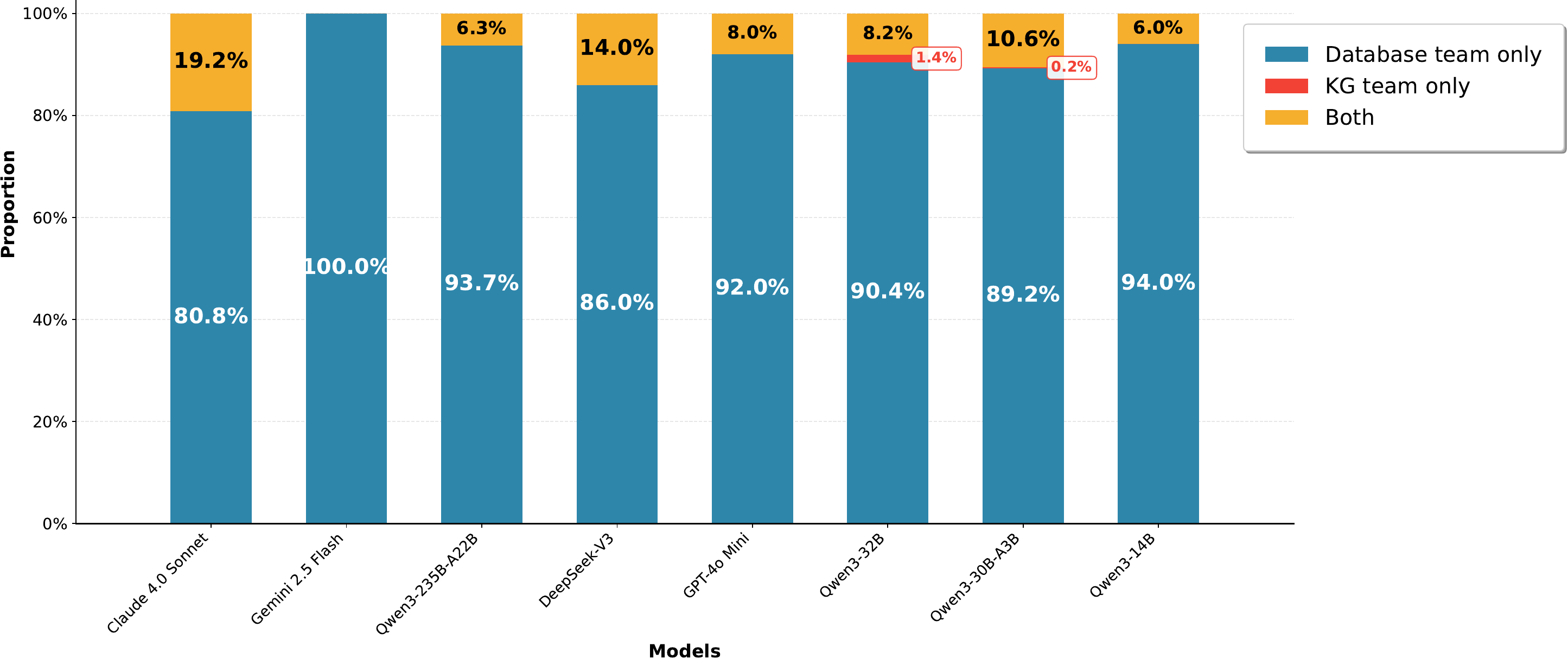}
		\caption{FeTaQA Dataset}
		\label{fig:tool_usage_fetaqa}
	\end{subfigure}
	
	\caption{Team invocation patterns across different models and datasets. The visualization demonstrates how task complexity influences tool selection strategies: (a) TabFact shows uniform Database Team preference due to binary classification nature, (b) WikiTQ exhibits moderate diversification with increased Knowledge Graph usage for complex retrieval tasks, and (c) FeTaQA displays highest combined team usage rates reflecting multi-hop reasoning requirements. Blue represents Database Team-only usage, red indicates Knowledge Graph Team-only usage, and orange represents combined team usage.}
	\label{fig:tool_usage_patterns}
\end{figure}

Our analysis demonstrates significant differences in how different models invoke the Database Team (SQL-based retrieval) and Knowledge Graph Team (Cypher-based retrieval) across the three benchmark datasets. The detailed analysis reveals distinct patterns based on task complexity and model characteristics (see Figure~\ref{fig:tool_usage_patterns}).

First, regarding dataset differences, the Data Leader's query strategies across different models are most uniform on the TabFact dataset, predominantly selecting Database Team-centered query approaches with minimal or no Knowledge Graph Team invocation. This occurs because TabFact tasks primarily involve binary classification verification, where the binary nature of fact verification favors direct SQL fact checking, and the relatively low task difficulty is evidenced by the generally high accuracy rates across different methods shown in Table~\ref{tab:comparison}. In contrast, on WikiTQ and FeTaQA datasets, models' query strategies become more diverse, with significantly increased Knowledge Graph Team invocation rates, and most models employ strategies that simultaneously invoke both Database and Knowledge Graph teams. Specifically, WikiTQ tasks require complex retrieval, matching, and computational analysis, exhibiting moderate strategy diversity with increased usage of the Knowledge Graph Team for relational reasoning. On FeTaQA, the task complexity and multi-hop reasoning characteristics lead models to prefer combined usage of both teams' tools, with combined usage rates reaching as high as 19.2\% (Claude 4.0 Sonnet), demonstrating that such free-form question answering problems rely more heavily on multi-source information collection and comprehensive analysis.

\clearpage

Across all three datasets, all models show a high preference for SQL-based database retrieval, while knowledge graph usage remains relatively limited across all models. We attribute this to the fact that existing TableQA tasks primarily examine structured querying rather than relational reasoning, with many tasks solvable directly through SQL. However, the lower invocation frequency of the Knowledge Graph Team does not indicate its weakness or ineffectiveness—rather, it reflects the complementary and specialized nature of graph-based reasoning. As demonstrated in our RQ4 analysis (Section~\ref{sec:rq4}), even when KG Team usage appears limited, its contribution leads to consistent performance improvements: 2.6\% on TabFact, 5.5\% on WikiTQ, and up to 7.8\% on FeTaQA ROUGE-2 metrics. The strategic value of the KG Team lies not in frequency of use, but in its precision for specific reasoning scenarios requiring semantic relationships and multi-hop logic. In practical applications, knowledge graph usage remains an important complement, particularly for complex relationship analysis that exceeds SQL's structural querying capabilities.

Regarding different model capabilities, the following characteristics emerge: Claude 4.0 Sonnet achieves the highest combined usage rate (19.2\%) on FeTaQA, indicating its superior planning capabilities for coordinating multiple teams, followed by DeepSeek-V3, with both models demonstrating diverse invocation strategies. Gemini 2.5 Flash appears most conservative, invoking only SQL across all datasets, suggesting its preference for simpler, more reliable problem-solving paths. In comparing dense versus MoE architectures, the MoE model Qwen3-30B-A3B shows higher combined tool usage rates across all three datasets, with average combined team usage (9.47\%) significantly exceeding that of the dense Qwen3-32B (5.67\%), suggesting that MoE architectures may adopt more exploratory retrieval strategies.

\begin{tcolorbox}[colback=gray!10, colframe=gray!50, title=\textbf{RQ3 Summary Answer}, boxrule=1pt]
Task complexity significantly influences team invocation patterns: TabFact favors single Database Team usage, while FeTaQA shows highest combined team usage (up to 19.2\% for Claude). All models prefer SQL-based retrieval overall, but advanced models like Claude 4.0 Sonnet demonstrate superior multi-team coordination capabilities. MoE architectures show more exploratory strategies with higher tool combination rates than dense models.The lower frequency of Knowledge Graph Team invocation reflects task-driven specialization rather than ineffectiveness—its contributions in practical deployment are primarily manifested in tasks centered on data associations and semantic relationships, where it provides unique reasoning capabilities.
\end{tcolorbox}

\subsection{RQ4: Impact of Knowledge Graph Team Removal on TableQA Performance}\label{sec:rq4}

To evaluate the necessity and effectiveness of the Knowledge Graph Team, we conducted comprehensive ablation experiments comparing performance without the Knowledge Graph Team versus the complete multi-agent framework across all eight models and three benchmark datasets. Table~\ref{tab:team_comparison} presents the systematic comparison between framework configurations excluding the KG Team (using only structured SQL querying) versus the complete framework with integrated SQL and Cypher-based retrieval approaches.

\begin{table}[htbp]
\centering
\caption{Performance comparison between framework without KG Team and complete framework}
\label{tab:team_comparison}
\resizebox{0.68\textwidth}{!}{%
\begin{tabular}{lllccc}
\toprule
\textbf{Dataset} & \textbf{Metric} & \textbf{Model} & \textbf{No KG Team} & \textbf{Complete Framework} & \textbf{Avg. Improvement} \\
\midrule
\multirow{8}{*}{TabFact} & \multirow{8}{*}{Accuracy} & Claude 4.0 Sonnet & 86.7\% & 90.8\% \textcolor{green}{($\uparrow$4.7\%)} & \multirow{8}{*}{5.5\%} \\
\cmidrule{3-5}
 &  & Gemini-2.5 Flash & 85.1\% & 87.9\% \textcolor{green}{($\uparrow$3.3\%)} & \\
\cmidrule{3-5}
 &  & Qwen3-235B-A22B & 85.6\% & 91.2\% \textcolor{green}{($\uparrow$6.5\%)} & \\
\cmidrule{3-5}
 &  & DeepSeek Chat & 82.0\% & 84.2\% \textcolor{green}{($\uparrow$2.7\%)} & \\
\cmidrule{3-5}
 &  & GPT-4o mini & 77.3\% & 79.3\% \textcolor{green}{($\uparrow$2.6\%)} & \\
\cmidrule{3-5}
 &  & Qwen3-32B & 67.7\% & 79.8\% \textcolor{green}{($\uparrow$17.9\%)} & \\
\cmidrule{3-5}
 &  & Qwen3-30B-A3B & 73.0\% & 75.0\% \textcolor{green}{($\uparrow$2.7\%)} & \\
\cmidrule{3-5}
 &  & Qwen3-14B & 73.0\% & 75.7\% \textcolor{green}{($\uparrow$3.7\%)} & \\
\midrule
\multirow{8}{*}{WikiTQ} & \multirow{8}{*}{Accuracy} & Claude 4.0 Sonnet & 78.2\% & 83.2\% \textcolor{green}{($\uparrow$6.4\%)} & \multirow{8}{*}{14.4\%} \\
\cmidrule{3-5}
 &  & Gemini-2.5 Flash & 74.0\% & 77.9\% \textcolor{green}{($\uparrow$5.3\%)} & \\
\cmidrule{3-5}
 &  & Qwen3-235B-A22B & 70.0\% & 73.2\% \textcolor{green}{($\uparrow$4.6\%)} & \\
\cmidrule{3-5}
 &  & DeepSeek Chat & 69.0\% & 73.6\% \textcolor{green}{($\uparrow$6.7\%)} & \\
\cmidrule{3-5}
 &  & GPT-4o mini & 64.9\% & 67.8\% \textcolor{green}{($\uparrow$4.4\%)} & \\
\cmidrule{3-5}
 &  & Qwen3-32B & 42.1\% & 61.0\% \textcolor{green}{($\uparrow$44.9\%)} & \\
\cmidrule{3-5}
 &  & Qwen3-30B-A3B & 45.8\% & 48.1\% \textcolor{green}{($\uparrow$5.0\%)} & \\
\cmidrule{3-5}
 &  & Qwen3-14B & 39.0\% & 53.7\% \textcolor{green}{($\uparrow$37.7\%)} & \\
\midrule
\multirow{24}{*}{FeTaQA} & \multirow{8}{*}{ROUGE-1 F} & Claude 4.0 Sonnet & 0.5963 & 0.6234 \textcolor{green}{($\uparrow$4.5\%)} & \multirow{8}{*}{12.8\%} \\
\cmidrule{3-5}
 &  & Gemini-2.5 Flash & 0.5829 & 0.6107 \textcolor{green}{($\uparrow$4.8\%)} & \\
\cmidrule{3-5}
 &  & Qwen3-235B-A22B & 0.552 & 0.5972 \textcolor{green}{($\uparrow$8.2\%)} & \\
\cmidrule{3-5}
 &  & DeepSeek Chat & 0.5161 & 0.5403 \textcolor{green}{($\uparrow$4.7\%)} & \\
\cmidrule{3-5}
 &  & GPT-4o mini & 0.5173 & 0.5576 \textcolor{green}{($\uparrow$7.8\%)} & \\
\cmidrule{3-5}
 &  & Qwen3-32B & 0.4209 & 0.5662 \textcolor{green}{($\uparrow$34.5\%)} & \\
\cmidrule{3-5}
 &  & Qwen3-30B-A3B & 0.4868 & 0.5063 \textcolor{green}{($\uparrow$4.0\%)} & \\
\cmidrule{3-5}
 &  & Qwen3-14B & 0.3954 & 0.5289 \textcolor{green}{($\uparrow$33.8\%)} & \\
\cmidrule{2-6}
 & \multirow{8}{*}{ROUGE-2 F} & Claude 4.0 Sonnet & 0.3675 & 0.3885 \textcolor{green}{($\uparrow$5.7\%)} & \multirow{8}{*}{17.1\%} \\
\cmidrule{3-5}
 &  & Gemini-2.5 Flash & 0.3634 & 0.3767 \textcolor{green}{($\uparrow$3.7\%)} & \\
\cmidrule{3-5}
 &  & Qwen3-235B-A22B & 0.3365 & 0.3607 \textcolor{green}{($\uparrow$7.2\%)} & \\
\cmidrule{3-5}
 &  & DeepSeek Chat & 0.2975 & 0.3075 \textcolor{green}{($\uparrow$3.4\%)} & \\
\cmidrule{3-5}
 &  & GPT-4o mini & 0.296 & 0.332 \textcolor{green}{($\uparrow$12.2\%)} & \\
\cmidrule{3-5}
 &  & Qwen3-32B & 0.218 & 0.3327 \textcolor{green}{($\uparrow$52.6\%)} & \\
\cmidrule{3-5}
 &  & Qwen3-30B-A3B & 0.2794 & 0.2899 \textcolor{green}{($\uparrow$3.8\%)} & \\
\cmidrule{3-5}
 &  & Qwen3-14B & 0.2042 & 0.3033 \textcolor{green}{($\uparrow$48.5\%)} & \\
\cmidrule{2-6}
 & \multirow{8}{*}{ROUGE-L F} & Claude 4.0 Sonnet & 0.4863 & 0.5067 \textcolor{green}{($\uparrow$4.2\%)} & \multirow{8}{*}{12.4\%} \\
\cmidrule{3-5}
 &  & Gemini-2.5 Flash & 0.4852 & 0.4995 \textcolor{green}{($\uparrow$2.9\%)} & \\
\cmidrule{3-5}
 &  & Qwen3-235B-A22B & 0.4522 & 0.4776 \textcolor{green}{($\uparrow$5.6\%)} & \\
\cmidrule{3-5}
 &  & DeepSeek Chat & 0.4193 & 0.4309 \textcolor{green}{($\uparrow$2.8\%)} & \\
\cmidrule{3-5}
 &  & GPT-4o mini & 0.4153 & 0.4555 \textcolor{green}{($\uparrow$9.7\%)} & \\
\cmidrule{3-5}
 &  & Qwen3-32B & 0.3394 & 0.4569 \textcolor{green}{($\uparrow$34.6\%)} & \\
\cmidrule{3-5}
 &  & Qwen3-30B-A3B & 0.3907 & 0.4091 \textcolor{green}{($\uparrow$4.7\%)} & \\
\cmidrule{3-5}
 &  & Qwen3-14B & 0.316 & 0.4253 \textcolor{green}{($\uparrow$34.6\%)} & \\
\bottomrule
\end{tabular}%
}
\end{table}

The comprehensive ablation study demonstrates that Knowledge Graph Team integration consistently improves performance across all datasets, with particularly pronounced benefits on multi-hop reasoning tasks. Performance gains scale systematically with task complexity: TabFact shows moderate improvements (5.5\% average) as binary fact verification primarily requires structured SQL queries, while WikiTQ (14.4\% average improvement) and FeTaQA (12.8\%-17.1\% improvements across metrics) reveal the substantial value of relational reasoning in complex retrieval and multi-hop inference tasks. Both WikiTQ and FeTaQA represent typical multi-hop question answering datasets, requiring semantic relationship understanding and cross-entity reasoning that traditional SQL-only approaches cannot adequately address.

Model capability tiers demonstrate distinct responses to multi-agent collaboration. Smaller parameter models (Qwen3-32B, Qwen3-14B) exhibit the most dramatic improvements under dual-team collaboration, achieving 44.9\% and 37.7\% accuracy gains on WikiTQ, and 52.6\% and 48.5\% ROUGE-2 improvements on FeTaQA respectively. These substantial gains on complex multi-hop datasets highlight how Knowledge Graph Team integration effectively compensates for smaller models' limitations in TableQA by enhancing their understanding and reasoning capabilities over complex relationships and semantic structures. Conversely, larger parameter models demonstrate more stable baseline performance with consistent but moderate collaboration benefits, reflecting different architectural adaptabilities to multi-agent coordination.

The relational reasoning capabilities provided by the Knowledge Graph Team through Cypher queries systematically address the fundamental limitations of SQL-only approaches. In scenarios requiring cross-row relationship analysis, semantic similarity judgment, and multi-hop reasoning paths, the KG Team's contributions become indispensable. This explains why collaborative effects are most pronounced in FeTaQA-type free-form question answering and WikiTQ-style complex retrieval tasks, both of which exemplify the multi-hop reasoning challenges that our framework is specifically designed to address. The automated knowledge graph construction enables discovery of implicit entity relationships and semantic connections that remain invisible to traditional structured query approaches, thereby supporting the sophisticated reasoning patterns required for advanced TableQA scenarios.

\begin{tcolorbox}[colback=gray!10, colframe=gray!50, title=\textbf{RQ4 Summary Answer}, boxrule=1pt]
The complete framework consistently outperforms configurations without the Knowledge Graph Team across all datasets, with improvements ranging from 5.5\% (TabFact) to 17.1\% (FeTaQA ROUGE-2). The comprehensive eight-model ablation study provides evidence that Knowledge Graph Team removal significantly degrades performance, particularly for semantic relationship analysis and cross-entity reasoning that characterize challenging multi-hop question answering scenarios. Performance losses scale systematically with task complexity, demonstrating the indispensable value of KG Team's relational reasoning capabilities in WikiTQ and FeTaQA datasets, both typical multi-hop reasoning benchmarks that our framework specifically targets.
\end{tcolorbox}

\subsection{RQ5: Impact of Team Collaboration Frequency on Performance}

\begin{figure}[H]
\centering
\includegraphics[width=0.85\textwidth]{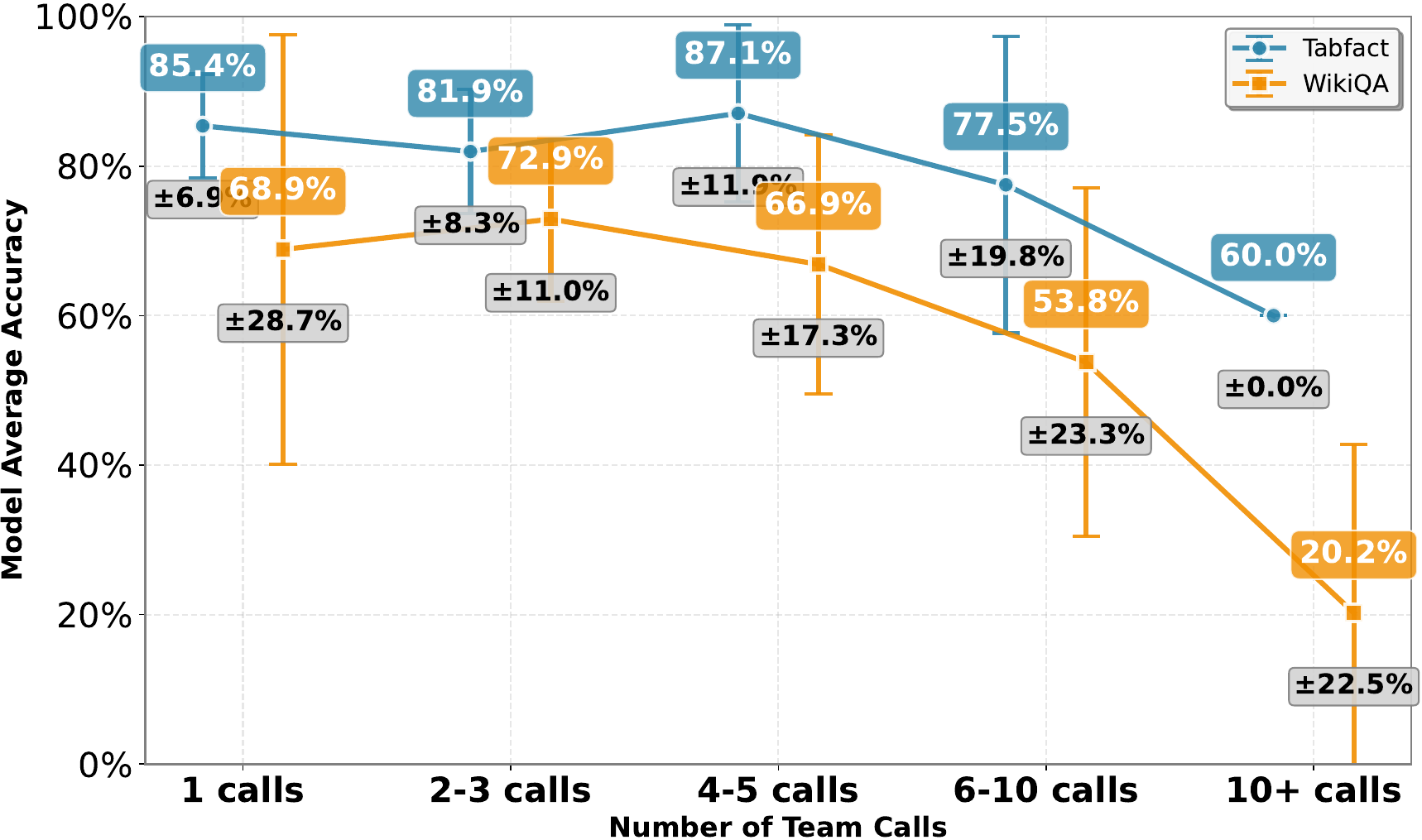}
\caption{Model average performance across different team call frequencies. Error bars represent standard deviation across all models. Blue line shows TabFact dataset results, orange line shows WikiTQ dataset results.}
\label{fig:performance_frequency}
\end{figure}

\begin{figure}[H]
\centering
\begin{subfigure}[b]{0.48\textwidth}
\centering
\includegraphics[width=\textwidth]{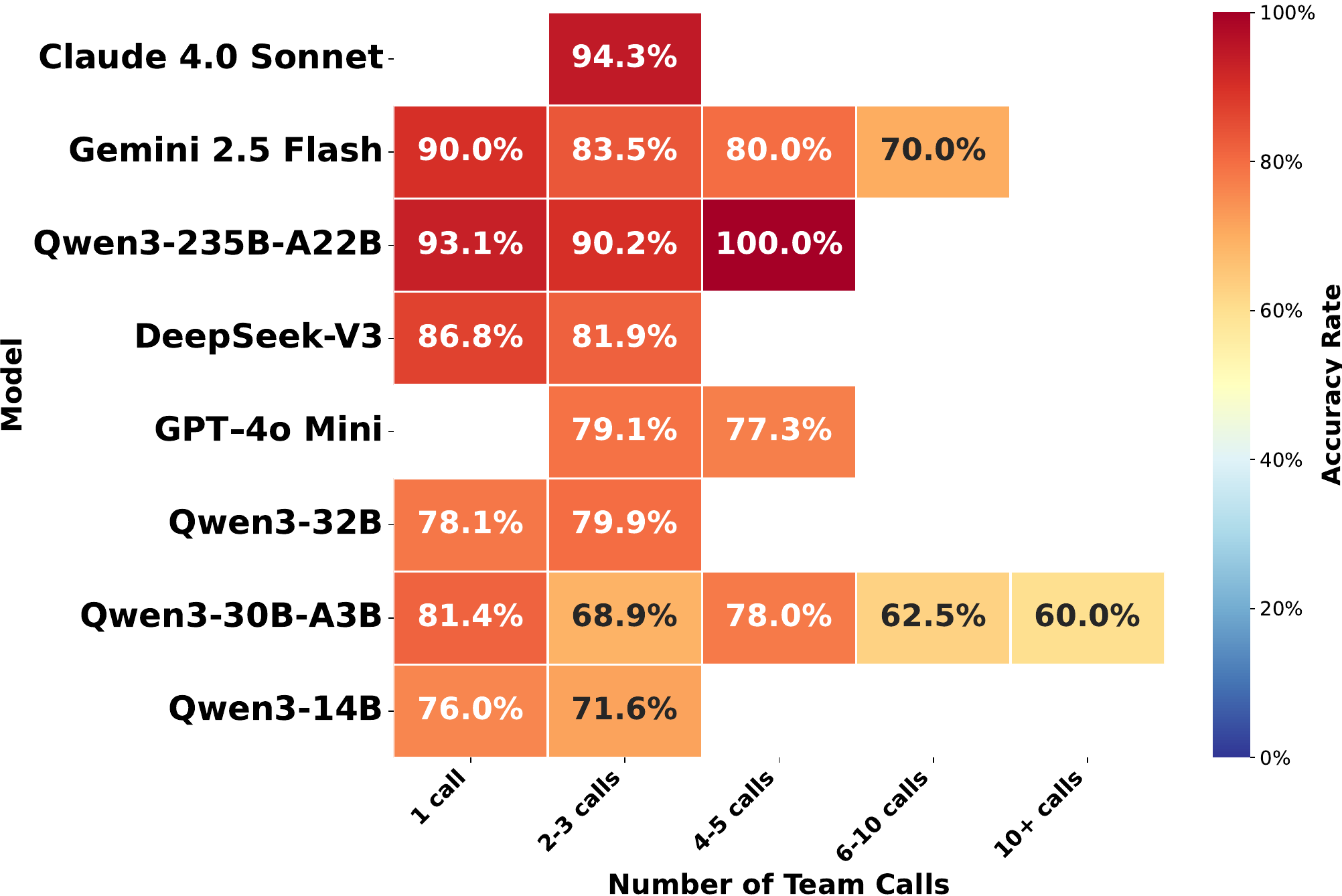}
\caption{TabFact Dataset}
\label{fig:tabfact_heatmap}
\end{subfigure}
\hfill
\begin{subfigure}[b]{0.48\textwidth}
\centering
\includegraphics[width=\textwidth]{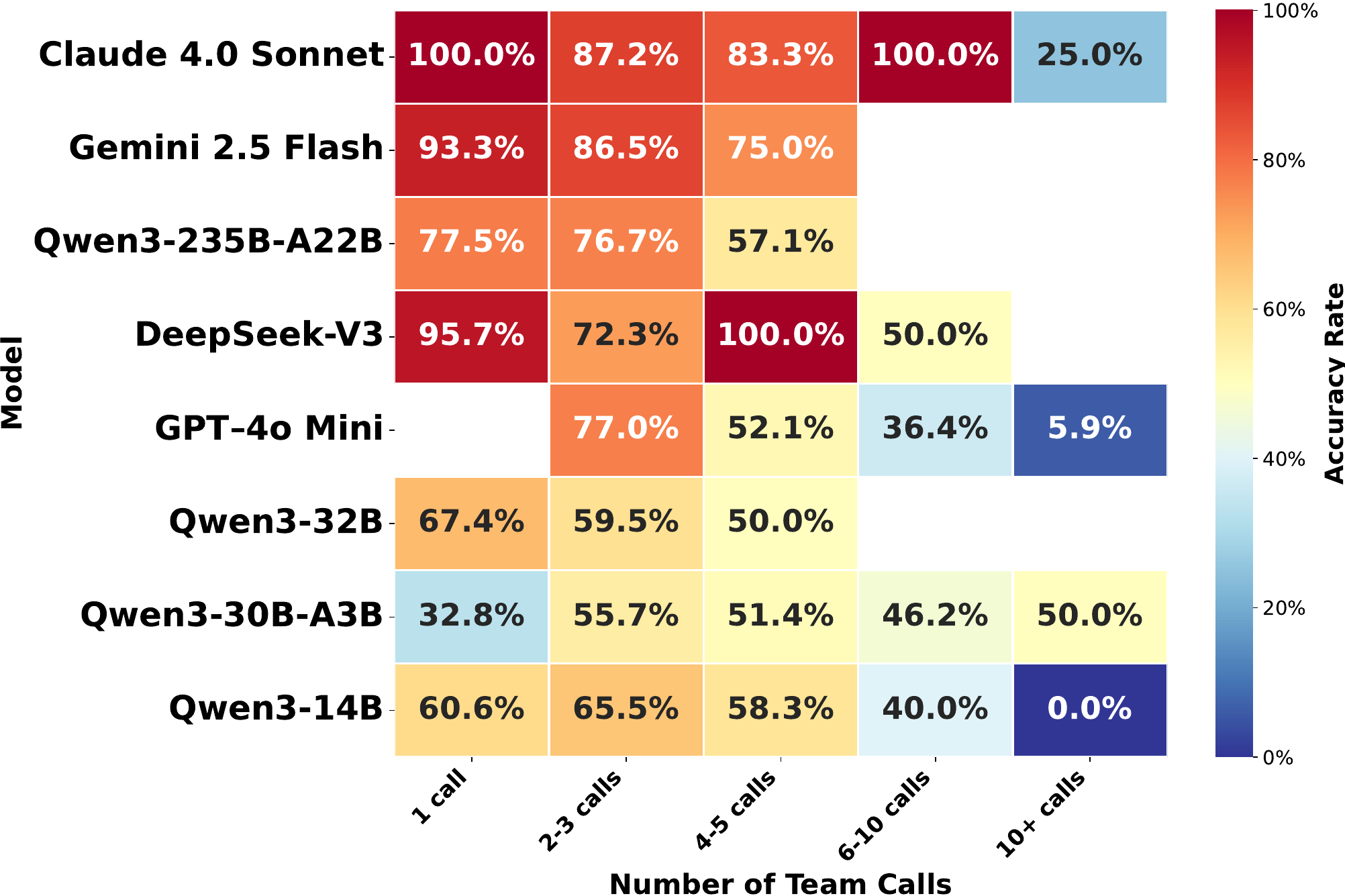}
\caption{WikiTQ Dataset}
\label{fig:wikitq_heatmap}
\end{subfigure}

\caption{Performance heatmaps showing model accuracy across different team collaboration call frequencies. The color intensity represents performance levels, with darker red indicating higher accuracy. Each row represents a different model, and each column represents a different call frequency range (1 call, 2--3 calls, 4--5 calls, 6--10 calls, 10+ calls).}
\label{fig:collaboration_heatmaps}
\end{figure}

To understand how collaboration frequency affects performance, we analyzed the relationship between team interaction frequency and task accuracy across all models on TabFact and WikiTQ datasets. We categorized interaction frequency into five intervals: single interaction (1 call), low-frequency (2--3 calls), medium-frequency (4--5 calls), high-frequency (6--10 calls), and extremely high-frequency (10+ calls). Our hypothesis posits that an optimal interaction frequency exists where performance is maximized, with degradation occurring at both extremes (see Figure~\ref{fig:performance_frequency} and Figure~\ref{fig:collaboration_heatmaps}).

Our analysis reveals a clear inverted U-shaped relationship between collaboration frequency and performance, with both datasets demonstrating similar patterns despite task complexity differences. Performance peaks in the low-frequency interaction range (1--3 calls) before gradually declining as interaction frequency increases.

The TabFact dataset exhibits optimal performance at 1 call (85.4\%) and 2--3 calls (81.9\%) with relatively small standard deviations (±6.9\% and ±8.3\%), indicating consistent performance across models for simple binary classification tasks. WikiTQ demonstrates peak performance at 4--5 calls (87.1\%), but experiences steeper decline, dropping to 20.2\% at 10+ calls with increased variance (±22.5\%). The large error bars at high-frequency interactions reflect the inherent performance differences among the eight models tested, which span different capability tiers and architectures. Additionally, high-frequency interaction scenarios (6+ calls) occur relatively infrequently during our experiments on these two datasets, leading to more dispersed data points as evidenced in Figure~\ref{fig:wikitq_heatmap}.

The heatmap analysis reveals distinct patterns based on model capabilities and task complexity. In TabFact (Figure~\ref{fig:tabfact_heatmap}), most models achieve optimal performance within 1--3 calls, with Claude 4.0 Sonnet reaching 94.3\% at 2--3 calls and Qwen3-235B achieving 100\% at 4--5 calls. This stability reflects the tolerance of verification tasks to interaction frequency. Conversely, WikiTQ (Figure~\ref{fig:wikitq_heatmap}) shows stronger early convergence, with Claude 4.0 Sonnet and DeepSeek-V3 achieving 100\% and 95.7\% respectively at single call, but rapid degradation with increased interactions.

Model capability tiers demonstrate distinct collaboration preferences. Top-tier models (e.g., Claude 4.0 Sonnet, Gemini 2.5 Flash) excel under low-frequency interactions, showcasing superior one-shot planning capabilities. Medium-scale models (Qwen3-32B, GPT-4o mini) benefit from moderate iteration (2--4 calls), while smaller models (Qwen3-14B) exhibit high volatility, declining from 65.5\% to 0\% on WikiTQ as interactions increase.

Most significantly, we observe a consistent downward trend when call counts exceed 6, with models experiencing performance drops on WikiTQ. GPT-4o mini's decline from 77.0\% (2--3 calls) to 5.9\% (10+ calls) exemplifies this pattern, suggesting that excessive collaboration leads to decision confusion, error accumulation, and disruption of reasoning chains. While individual model performance varies significantly at high frequencies due to their diverse capabilities, the overall degradation trend remains consistent across different model types.

\begin{tcolorbox}[colback=gray!10, colframe=gray!50, title=\textbf{RQ5 Summary Answer}, boxrule=1pt]
An optimal interaction frequency exists for team collaboration, with performance peaking at 1--3 calls and declining thereafter. Top-tier models demonstrate superior early convergence capabilities, while excessive collaboration (6+ calls) leads to severe performance degradation, particularly on complex tasks. The relationship between model capability and optimal collaboration frequency provides crucial insights for designing adaptive stopping mechanisms in multi-agent systems.
\end{tcolorbox}

\clearpage
\section{Theoretical and Practical Implications}

Our proposed LLM-based multi-agent DataFactory for TableQA provides insights into theoretical understanding and practical applications in the intersection of artificial intelligence, database systems, and natural language processing. Our framework transforms traditional single-agent approaches into a specialized team collaboration paradigm, contributing to multi-agent systems research while bringing tangible value to tabular data analysis applications.

\subsection{Theoretical Contributions}

From a theoretical perspective, our work expands the understanding of large language model multi-agent collaboration by demonstrating how specialized team architectures can overcome fundamental limitations of individual LLMs. The tripartite collaborative design—comprising a Data Leader employing the ReAct paradigm alongside Database Team and Knowledge Graph Team—provides insights into task decomposition and coordination in complex reasoning scenarios. A key theoretical contribution lies in our natural language-based consultation mechanisms that replace rigid workflow execution with flexible inter-agent communication, enabling adaptive strategy adjustment and knowledge sharing during complex reasoning processes. This architectural innovation addresses the theoretical challenge of balancing ``specialization'' and ``coordination,'' demonstrating that complementary advantages between structured data processing (SQL-based) and relational knowledge representation (graph-based) can be systematically integrated through intelligent orchestration and natural semantic interaction.

The automated ``data-to-knowledge graph'' transformation algorithm provides a methodological approach to knowledge representation learning. We formalized the mapping function $\mathcal{T}: \mathcal{D} \times \mathcal{S} \times \mathcal{R} \rightarrow \mathcal{G}$ and implemented entity construction and relationship discovery mechanisms, providing a systematic approach for transforming structured tabular data into semantic knowledge graphs. This approach extends beyond TableQA and offers insights for other domains requiring transformation of structured data into relational representations to enhance reasoning capabilities.

Furthermore, our context-enhanced prompting strategies for SQL and Cypher generation provide insights into how to integrate external knowledge, historical patterns, and domain information systems to reduce large model hallucinations. By combining RAG with specialized prompt engineering, we reduce hallucinations and improve query generation accuracy. Additionally, our experimental results reveal an inverted U-shaped relationship between collaboration frequency and performance, providing empirical evidence about effective agent interaction patterns in multi-agent systems.

\subsection{Practical Significance and Platform Integration}

At the practical level, DataFactory breaks down critical barriers in enterprise data analysis through natural language interfaces, enabling non-technical users to access complex tabular data. The framework combines the Database Team (responsible for structured queries) with the Knowledge Graph Team (responsible for relational reasoning), allowing users to simultaneously explore numerical patterns and semantic associations without requiring professional SQL or data science skills.

The platform implements an end-to-end integrated workflow encompassing data ingestion, knowledge graph construction, query processing, and result visualization. The web interface integrates multiple functional modules where specialized agents collaborate to automatically complete SQL and Cypher queries, data analysis, and visualization, thereby greatly reducing the threshold for data analysis and improving decision-making efficiency. 

In practical deployment, business analysts can propose complex queries involving cross-table relationships, temporal patterns, and multi-hop reasoning. The system returns structured analytical results and knowledge graph visualizations, significantly accelerating organizational learning cycles and reducing dependence on specialized data teams. The platform adopts a modular architecture, ensuring scalability and customizability across enterprise environments. It supports dual-mode deployment—local deployment based on open-source models and cloud deployment based on commercial APIs—to meet different resource and privacy requirements. 

Beyond quantitative results, the platform generates explanatory insights, enhancing transparency and credibility of automated analysis. This enables users to not only obtain answers but also understand how the Database Team and Knowledge Graph Team collaboratively reason to reach conclusions. The complete workflow demonstrates structured data analysis, knowledge relationship mining, and multi-agent collaborative decision-making (see Appendix for interface demonstrations).
\section{Conclusion and Future Work}

This paper presents an LLM-based multi-agent DataFactory framework that overcomes key limitations of single-agent TableQA approaches through systematic task decomposition and specialized team coordination. The framework contributes three technical innovations: (1) a ReAct-based Data Leader that orchestrates intelligent task delegation between Database and Knowledge Graph Teams, (2) automated data-to-knowledge graph transformation algorithms that capture both structural and semantic relationships, and (3) context-enhanced engineering strategies that improve query generation accuracy for both SQL and Cypher operations. Experimental validation across three benchmark datasets (TabFact, WikiTQ, FeTaQA) and eight models from five providers demonstrates consistent performance improvements, with our method achieving average improvements of 20.2\% on TabFact and 23.9\% on WikiTQ compared to baseline methods, and Cohen's d values greater than 1 indicating statistically significant enhancement effects. The collaborative coordination between specialized teams outperforms single-team approaches by 5.5\% on TabFact, 14.4\% on WikiTQ, and up to 17.1\% on FeTaQA ROUGE-2 metrics, with effectiveness scaling appropriately across different model capabilities and task complexities.

Future research should address several key areas: computational efficiency optimization for large-scale deployment, more systematic scalability evaluation under streaming and multi-tenant workloads, adaptive stopping mechanisms to prevent excessive multi-agent interactions, and enhanced robustness of automated knowledge graph construction. Another promising direction is to expose DataFactory as a back-end reasoning service for a broader class of conversational agents, including task-oriented and multi-modal dialogue systems. These improvements would strengthen the framework's practical applicability while preserving its core advantages in complex multi-hop reasoning tasks.

\appendix

\captionsetup[figure]{
  font={sf,small},
  labelfont={bf,color=black},
  labelsep=colon,
  singlelinecheck=true,
  skip=6pt
}

\section{Platform Interface Demonstrations}

The DataFactory framework has been implemented as a web-based platform that provides intuitive interfaces for data exploration and multi-agent collaboration. The following figures demonstrate the key functionalities of the three main components.

\vspace{1em}
\noindent\begin{minipage}{\textwidth}
    \centering
    \includegraphics[width=0.9\textwidth]{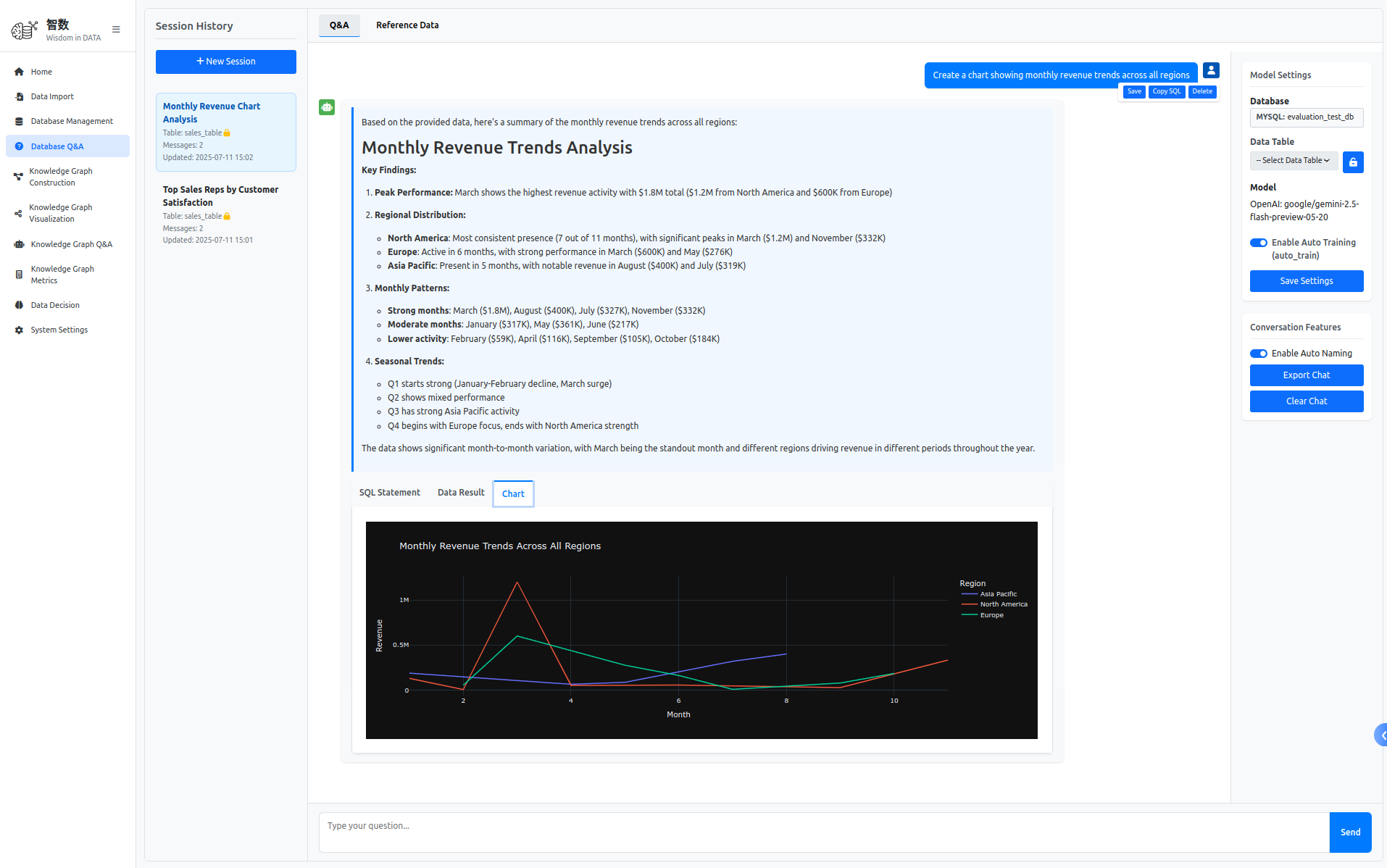}
    \captionof{figure}{Database Team Q\&A Interface: Users propose multi-region monthly revenue trend analysis requests through natural language, with the system automatically generating SQL and visualizing results.}
    \label{fig:data_qa_chart}
\end{minipage}

\noindent\begin{minipage}{\textwidth}
    \centering
    \includegraphics[width=0.9\textwidth]{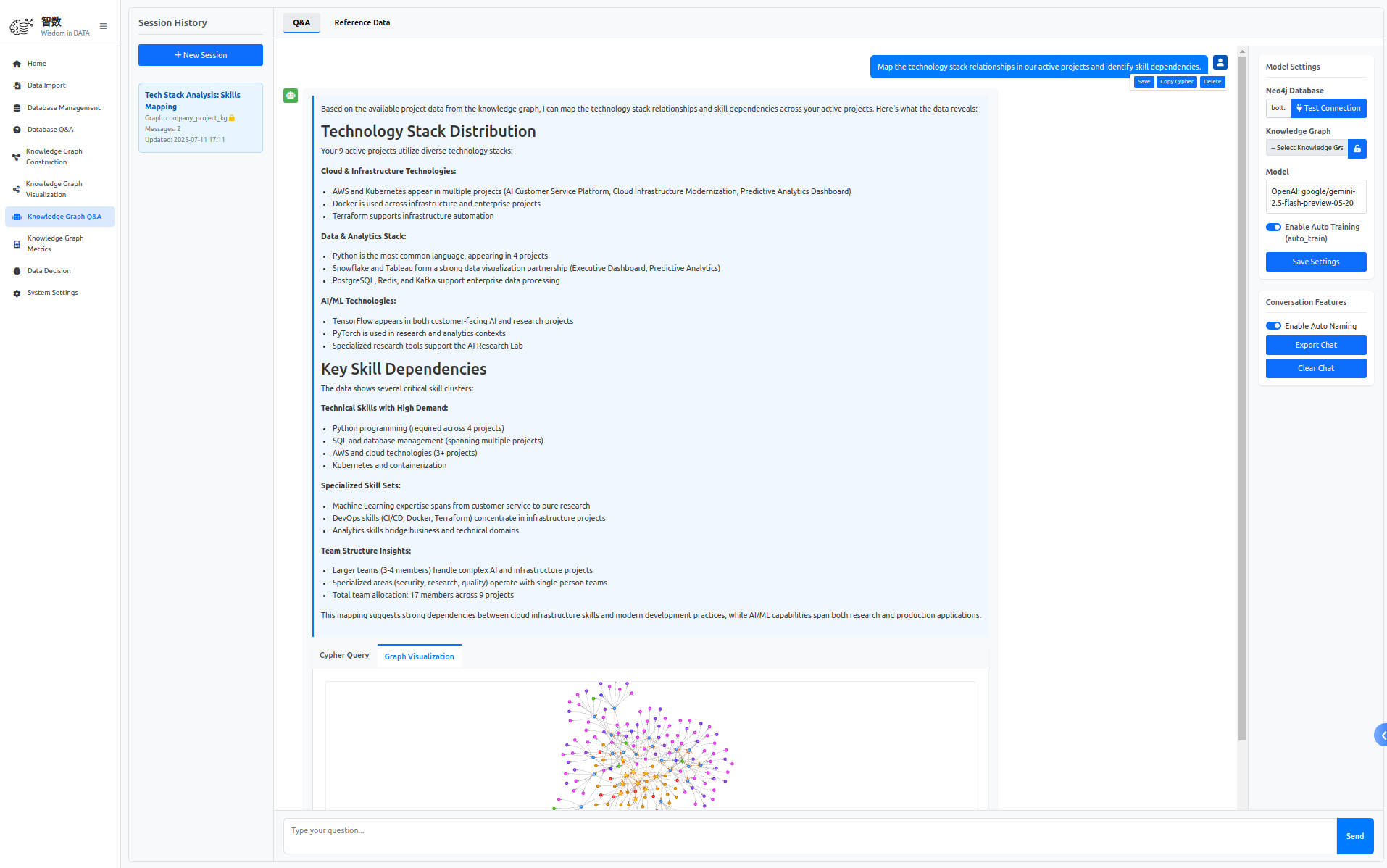}
    \captionof{figure}{Knowledge Graph Team Q\&A Interface: Users propose technology stack relationship and skill dependency analysis requests, with the system automatically generating Cypher queries and visualizing knowledge graph subgraphs.}
    \label{fig:kg_qa_subgraph}
\end{minipage}

\vspace{1em}
\noindent\begin{minipage}{\textwidth}
    \centering
    \includegraphics[width=0.9\textwidth]{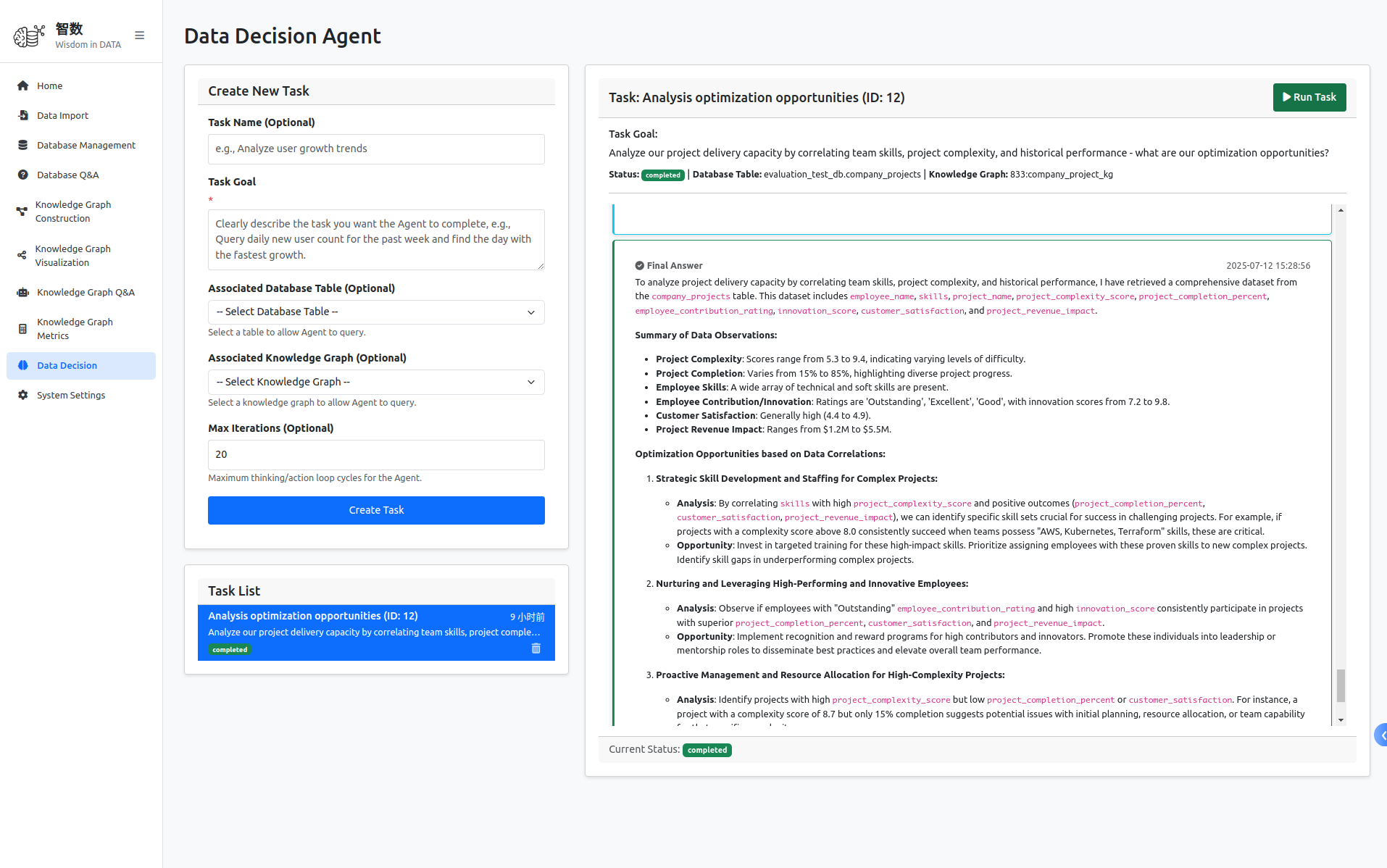}
    \captionof{figure}{DataFactory Decision Analysis Interface: Under multi-agent collaboration, the Leader coordinates Database and Knowledge Graph teams for collaborative analysis and provides decision recommendations.}
    \label{fig:data_factory_decision}
\end{minipage}
\vspace{1em}
\FloatBarrier  

\section*{Acknowledgments}

We would like to express our sincere gratitude to the National Defense Science and Technology Key Laboratory Fund of China (Grant No. 6142006220506) for their financial support of this research. We also thank the anonymous reviewers for their valuable comments and suggestions that have significantly improved the quality of this paper.

\printcredits

\bibliographystyle{cas-model2-names}

\bibliography{cas-refs-cleaned}





\end{document}